\documentclass[conference]{IEEEtran}
\IEEEoverridecommandlockouts
\usepackage{cite}
\usepackage{amsmath,amssymb,amsfonts}
\usepackage[pdftex]{graphicx}
\usepackage{textcomp}
\usepackage{xcolor}

\usepackage{booktabs} 

\usepackage[english]{babel}
\usepackage{moresize}
\usepackage{amsmath}

\usepackage{balance}
\usepackage{comment}
\usepackage{paralist}
\usepackage{multirow}

\usepackage{filecontents}

\usepackage{flushend}
\usepackage[english]{babel}
\usepackage[latin1]{inputenc}
\usepackage{mathrsfs}
\usepackage{graphicx}
\usepackage{amssymb}
\usepackage{url}
\usepackage{subfigure}
\usepackage{amsmath}
\usepackage{enumitem}
\usepackage[linesnumbered,algoruled,boxed,lined]{algorithm2e}
\usepackage{adjustbox}
\usepackage{amssymb}
\usepackage{times}
\usepackage{hyperref}

\usepackage{pgfplots}
\usetikzlibrary{pgfplots.dateplot}

\usepackage{filecontents}
\definecolor{tblue}{RGB}{31,119,180}
\definecolor{torange}{RGB}{255,127,14}
\definecolor{tgreen}{RGB}{44,160,44}
\definecolor{tred}{RGB}{214,39,40}
\definecolor{tpurple}{RGB}{148,103,189}

\usepackage{filecontents}

\newcommand{\hide}[1]{} 

\newcommand{\etal}{\textit{et al}.}

\newcommand{\ie}{\textit{i}.\textit{e}.}
\newcommand{\eg}{\textit{e}.\textit{g}.}

\def\BibTeX{{\rm B\kern-.05em{\sc i\kern-.025em b}\kern-.08em
    T\kern-.1667em\lower.7ex\hbox{E}\kern-.125emX}}
    
\begin{document}


\title{Spatial-Temporal Hypergraph Self-Supervised Learning for Crime Prediction

\thanks{\textbf{*Corresponding author: Chao Huang.}}}

\def\model{ST-HSL}
\def\full{Spatial-Temporal Hypergraph Self-Supervised Learning}



\author{\IEEEauthorblockN{Zhonghang Li$^1$, Chao Huang$^{2,3*}$, Lianghao Xia$^{2,3}$, Yong Xu$^1$, Jian Pei$^4$}
\IEEEauthorblockA{South China University of Technology$^1$, Simon Fraser University$^4$ \\
Department of Computer Science$^2$, Musketeers Foundation Institute of Data Science$^3$, University of Hong Kong\\
cszhonghang.li@mail.scut.edu.cn, chaohuang75@gmail.com, aka\_xia@foxmail.com, yxu@scut.edu.cn, jpei@cs.sfu.ca\
}}



\maketitle
\begin{abstract}
Crime has become a major concern in many cities, which calls for the rising demand for timely predicting citywide crime occurrence. Accurate crime prediction results are vital for the beforehand decision-making of government to alleviate the increasing concern about the public safety. While many efforts have been devoted to proposing various spatial-temporal forecasting techniques to explore dependence across locations and time periods, most of them follow a supervised learning manner, which limits their spatial-temporal representation ability on sparse crime data. Inspired by the recent success in self-supervised learning, this work proposes a \underline{S}patial-\underline{T}emporal \underline{H}ypergraph \underline{S}elf-Supervised \underline{L}earning framework (\model) to tackle the label scarcity issue in crime prediction. Specifically, we propose the cross-region hypergraph structure learning to encode region-wise crime dependency under the entire urban space. Furthermore, we design the dual-stage self-supervised learning paradigm, to not only jointly capture local- and global-level spatial-temporal crime patterns, but also supplement the sparse crime representation by augmenting region self-discrimination. We perform extensive experiments on two real-life crime datasets. Evaluation results show that our \model\ significantly outperforms state-of-the-art baselines. Further analysis provides insights into the superiority of our \model\ method in the representation of spatial-temporal crime patterns. The implementation code is available at https://github.com/LZH-YS1998/STHSL.\\\vspace{-0.05in}
\end{abstract}


\section{Introduction}
\label{sec:intro}
Crime prevention has become a critical issue for many cities, which endangers the public safety seriously~\cite{chen2004crime,maguire2017crime}. The increase of crime activities not only affects individuals, but also the businesses and societies. According to the research statistic~\cite{miller2021incidence}, the cost of crimes is over \$2.5 trillion in USA 2017. Hence, effective crime prevention strategies can bring substantial economic benefits and reducing crime-relevant cost (\eg, business loss, violence with injury, and death). With the advancement of data positioning and acquisition techniques in urban sensing, there is a rising demand for accurately forecasting future crime occurrence~\cite{wang2016crime}. Such prediction results can be beneficial for the policy-making of government by proposing effective solutions (\eg, police dispatch) beforehand, so as to reduces the crime rate~\cite{wang2017non}.

Predicting urban crimes of each geographical region in a city, however, is very challenging, due to the complex spatial and temporal crime patterns vary by locations and time periods. Among various spatial-temporal prediction methods, deep learning approaches stand out owing to the strong feature representation ability of neural network architectures~\cite{wu2021quantifying}. There exist many recently developed forecasting models focusing on encoding temporal dependency and region-wise spatial dynamics. To be specific, attentive models propose to fuse spatial-temporal information with various attention mechanism, such as recurrent attentive network in DeepCrime~\cite{huang2018deepcrime}, the periodically shifted attention in STDN~\cite{yao2019revisiting} and the multi-level attention network in GeoMAN~\cite{liang2018geoman}. Motivated by the strength of graph neural networks, another relevant research line is to explore spatial-temporal graph structure for making predictions, \eg, graph convolution-based method STGCN~\cite{yubingspatio}, and attentional graph message passing schemes in GMAN~\cite{zheng2020gman} and ST-MetaNet~\cite{pan2019urban}.

\begin{figure}[t]
    \centering
    \subfigure[New York City]{
        \centering
        \includegraphics[width=0.42\columnwidth]{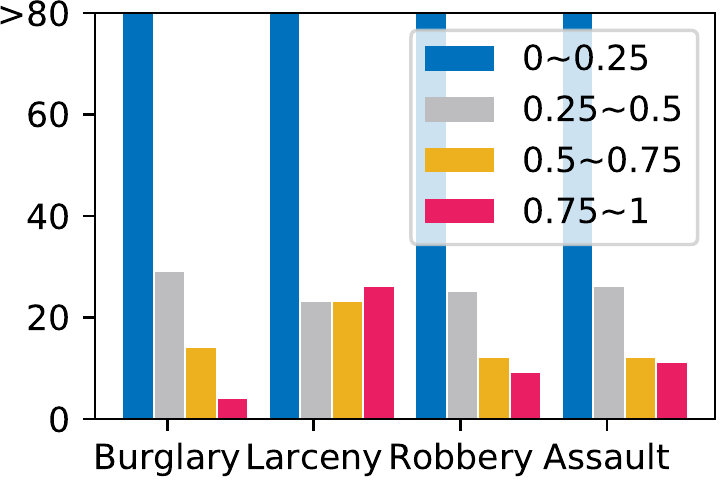}
    }
    \subfigure[Chicago]{
        \centering
        \includegraphics[width=0.42\columnwidth]{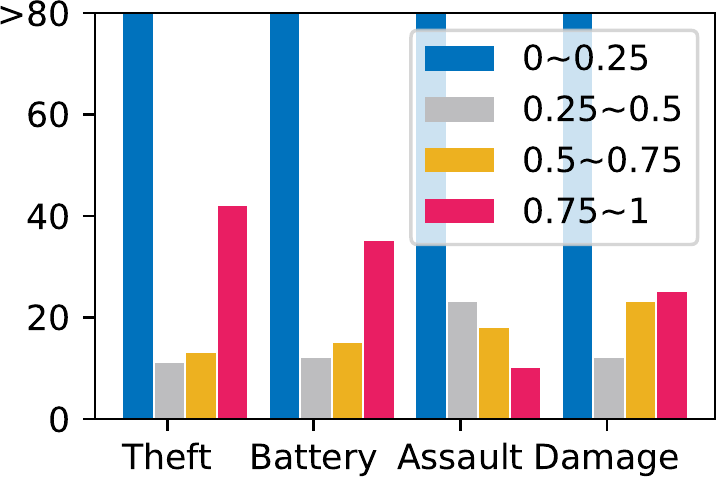}
    }
    \vspace{-0.05in}
    \caption{Distribution of crime sequence density degrees (\ie, the ratio of non-zero elements with crime occurrence) of regions at NYC and Chicago.}
    \vspace{-0.20in}
    \label{fig:sparse_dis}
\end{figure}

\begin{figure*}[t]
    \centering
    \subfigure[Burglary]{
        \centering
        \includegraphics[width=0.4\columnwidth]{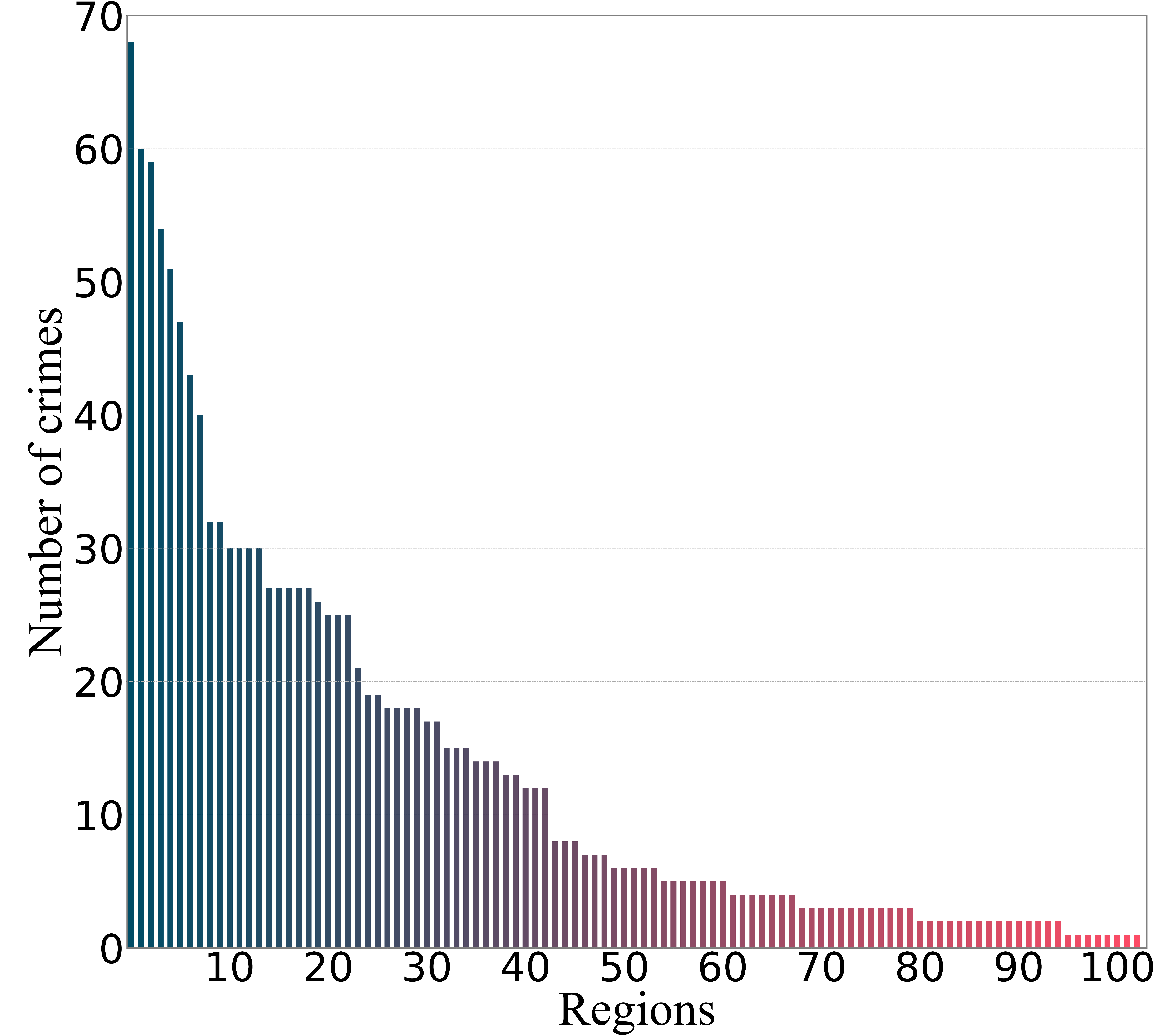}
    }
    \subfigure[Larceny]{
        \centering
        \includegraphics[width=0.4\columnwidth]{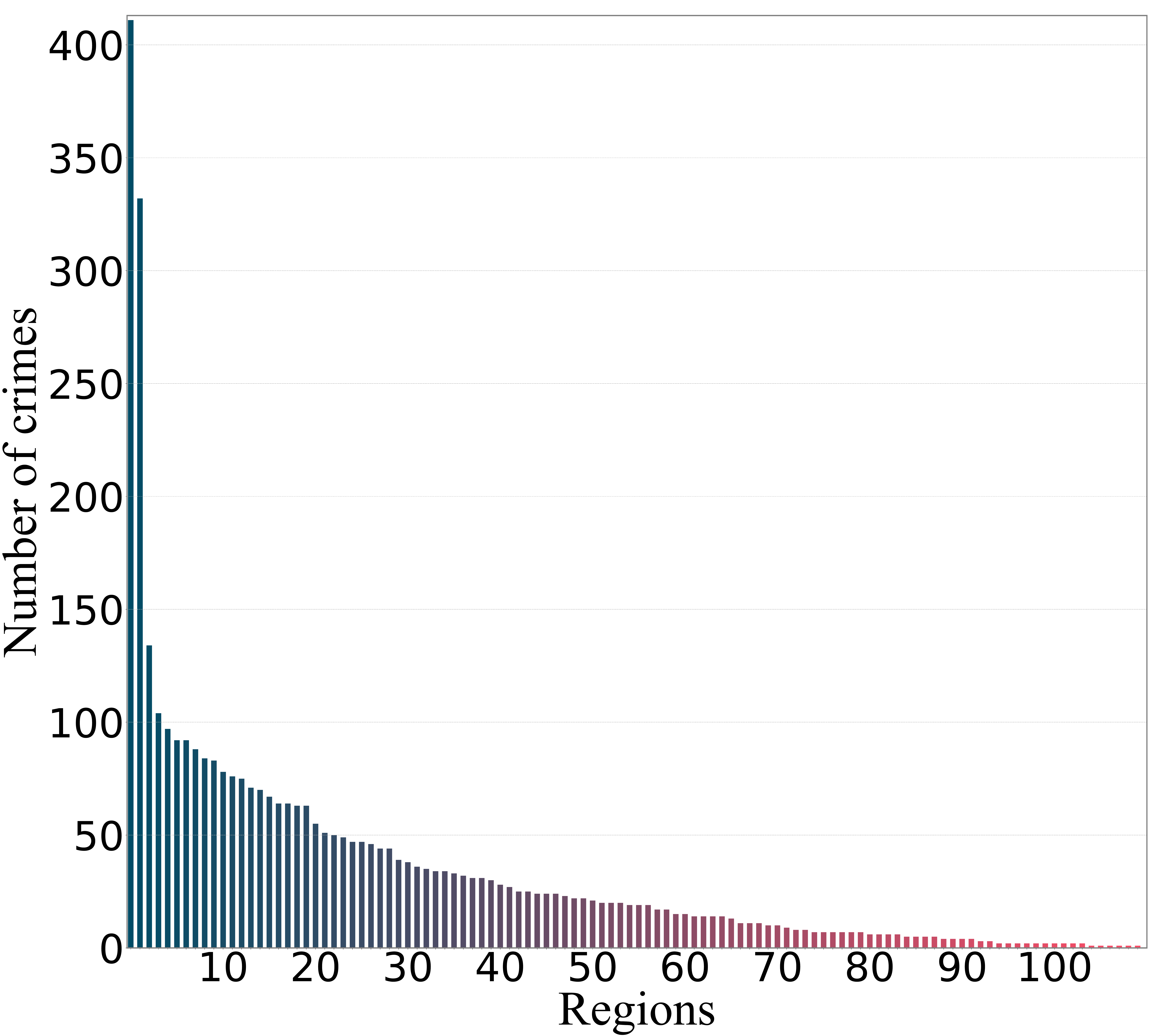}
    }
    \subfigure[Robbery]{
        \centering
        \includegraphics[width=0.4\columnwidth]{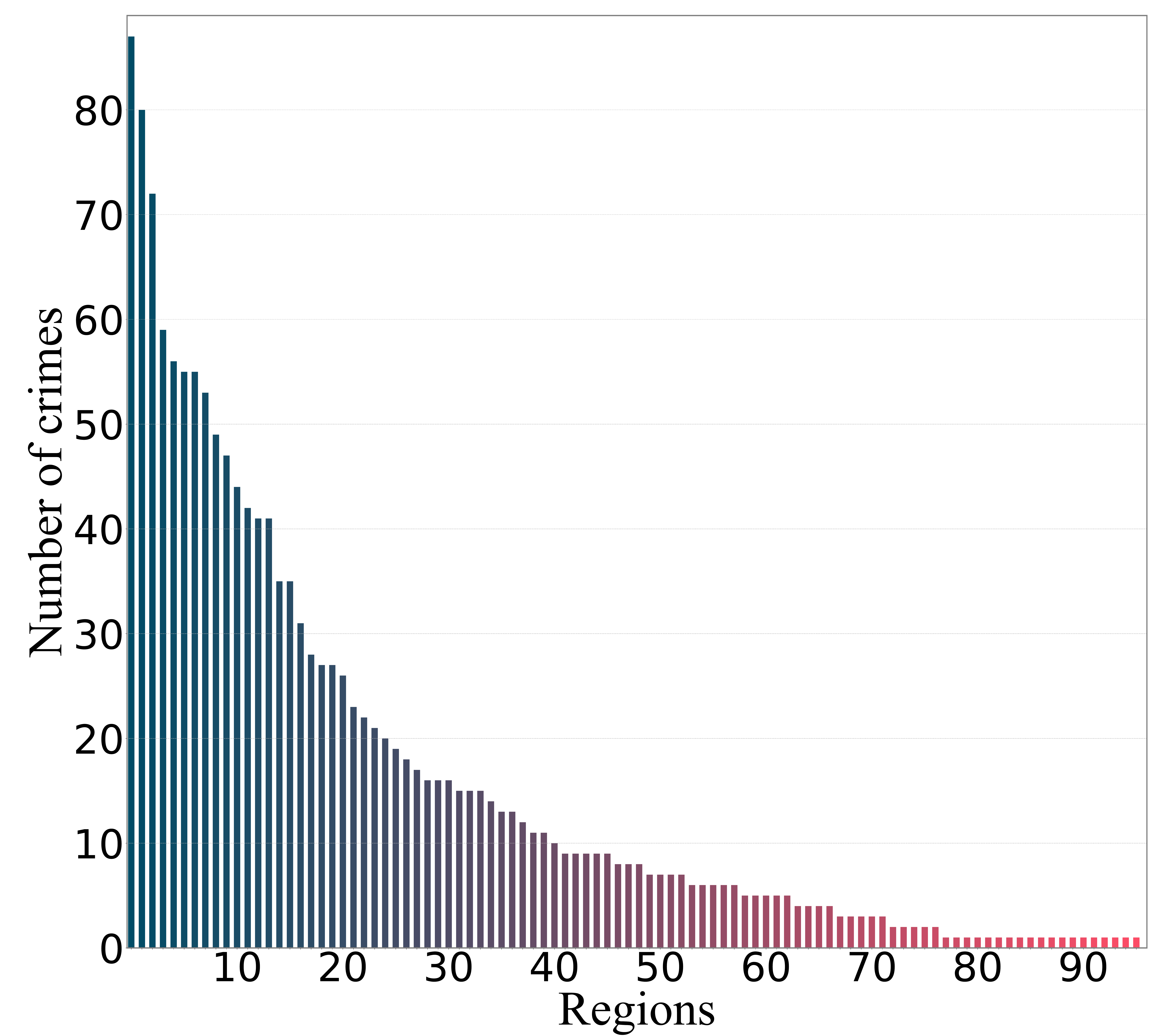}
    }
    \subfigure[Assault]{
        \centering
        \includegraphics[width=0.4\columnwidth]{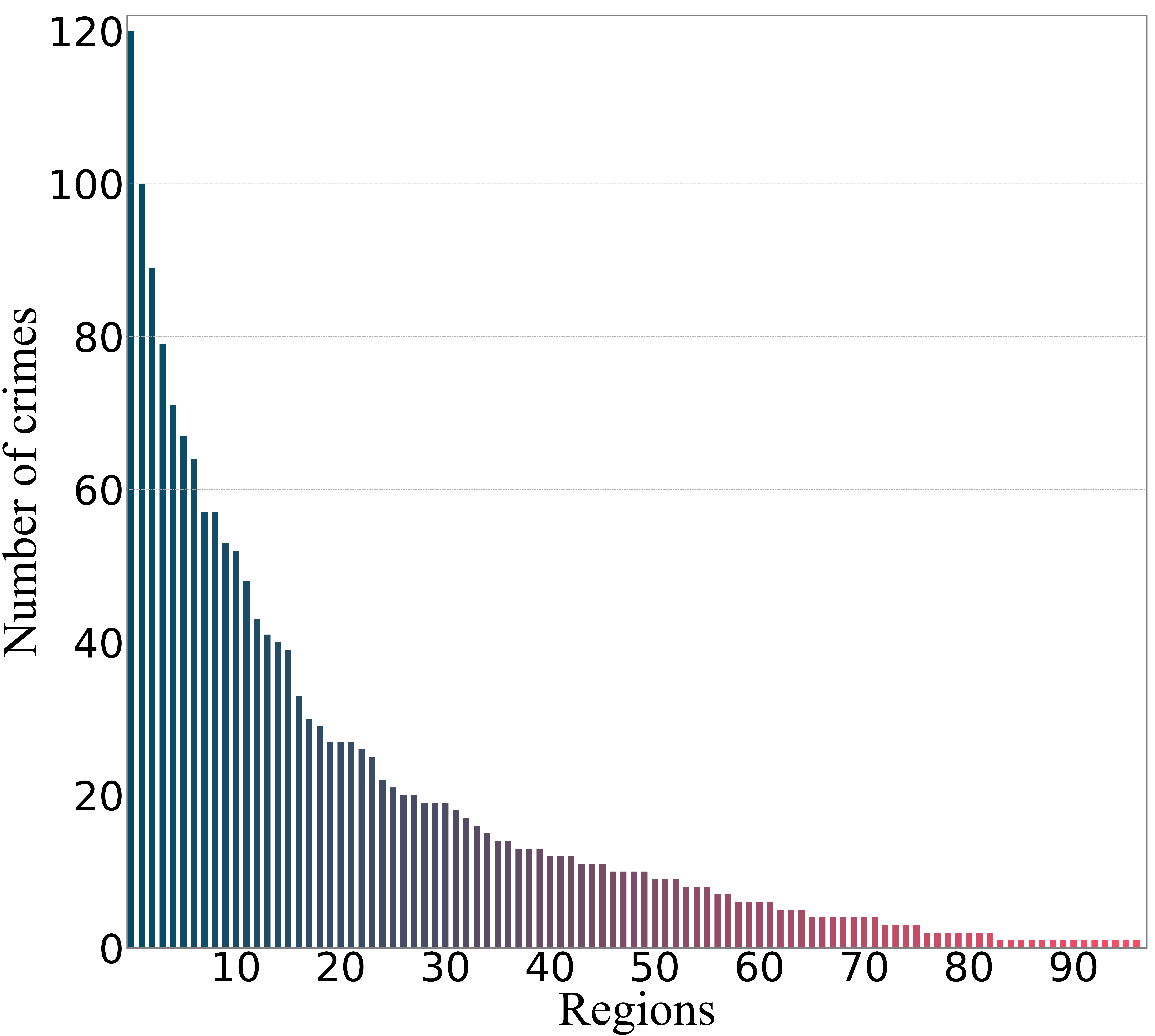}
    }
    \vspace{-0.05in}
    \caption{Skewed distribution of crime occurrence at geographical regions reported in New York City on September 2015.}
    \vspace{-0.1in}
    \label{fig:data_dist}
\end{figure*}

However, the aforementioned work has thus far focused on spatial-temporal representation. Despite their effectiveness, most of existing spatial-temporal prediction methods suffer from the limitation in predicting urban crimes with the sparse supervision signals. The crime prediction problem present unique challenges as follows:

\begin{itemize}[leftmargin=*]

\item \textbf{Sparse Supervision Signal}. Current spatial-temporal prediction models approach the spatial-temporal prediction task under a supervised learning framework, which requires sufficient supervision labels to learn quality representations. However, the urban crime data of each fine-grained region is extremely sparse, compared with the entire urban space~\cite{zhao2016multi,2020hierarchically}. For example, crime events can only occur at a small number of geographical areas across the whole city. We present the distribution of crime sequence density degrees of geographical regions in New York City and Chicago in Figure~\ref{fig:sparse_dis}. In particular, the region-specific density degree is generated by calculating the percentage of days over the two-year period with crime occurrences. From our statistical information, we can observe that the crime sequences of most regions have low density degrees (\eg, $[0,0.25]$). Such sparse supervision crime information make current deep models easily towards overfitting with the generated sub-optimal spatial-temporal relationship embeddings. \\\vspace{-0.10in}

\item \textbf{Skewed Crime Data Distribution}. The crime occurrence of different regions in a city often exhibit a power-law distribution, as shown in Figure~\ref{fig:data_dist}. Specifically, we show the skewed distribution of crime occurrence with different types at different regions in New York City on September 2015. The y-axis represents the number of crime cases reported from the time period, and x-axis denotes the region index. For better presentation for the long-tail crime distribution, we sort the region index (x-axis) in terms of the number of crime cases. Therefore, each bar in those figures represent the number of happened crimes with a certain crime category at a specific region. When learning spatial dependencies among geographical locations, neural network models are easily affected by the regions with more frequent crime occurrence, and will sacrifice the prediction performance of low-degree regions. The neighborhood information aggregation mechanism in state-of-the-art GNN methods will enlarge the effect of skewed data distribution. Therefore, it is a necessity to enable the robust spatial-temporal representation under imbalanced crime data.\\\vspace{-0.1in}


\end{itemize}

\noindent \textbf{Presented work.} In light of the aforementioned challenge and limitation, we proposed a spatial-temporal self-supervised learning framework for urban crime prediction, which we refer to as \underline{S}patial-\underline{T}emporal \underline{H}ypergraph \underline{S}elf-Supervised \underline{L}earning, \model\ for brevity. In \model, we first develop a multi-view spatial-temporal convolution network to encode the local dependency among nearby regions and time periods, as well as the implicit type-wise crime correlations. Then, we explore the potentials of integrating hypergraph structure learning with graph neural architecture, to capture global cross-region crime dependency. Technically, we generalize the concept of hyperedge by making it as intermediate hubs to interact with different regions. By designing hypergraph message passing schema, \model\ aggregates regions' crime patterns through multiple channels, and yields comprehensive crime representations with the preservation of spatial-temporal context.

We further design a novel spatial-temporal hypergraph contrastive learning paradigm which enables the local and global relation encoder to collaboratively supervise each other, to perform robust spatial-temporal representation with sparse crime data. \model\ effectively constructs augmentation from unlabeled crime data, and enhances the discrimination ability of our model in differentiating spatial-temporal crime patterns of different regions and time periods under data scarcity. To supercharge \model\ to inject global crime context, hypergraph infomax network is introduced to achieve the agreement between the local- and global-level representations.

We summarize key contributions of this paper as follows:
\begin{itemize}[leftmargin=*]

\item We propose novel crime prediction model \model\ that addresses the issue of sparse supervision signals and skewed crime data distribution, by unifying hypergraph dependency modeling with self-supervision learning for spatial-temporal crime representation.\\\vspace{-0.15in}

\item \model\ innovatively endow local- and global-level spatial-temporal encoder to perform cross-view collaborative supervision under a hypergraph contrastive learning paradigm. Moreover, a hypergraph infomax network is designed to further enhance the contrastive schema with the modeling of global crime spatial-temporal context.\\\vspace{-0.15in}


\item We perform extensive experiments on two real-world crime datasets, with detailed analyses for the effectiveness of our \model\ framework as compared to 15 spatial-temporal forecasting methods from various aspects. Furthermore, detailed model ablation study demonstrates the rationality of key components in our proposed new framework.


\end{itemize}

\section{Preliminaries}
\label{sec:model}


In this section, we first introduce preliminaries and key definitions in this work, and then we formally present our studied problem of urban crime prediction. Table~\ref{tab:notations} presents the frequently used notations throughout this paper. \\\vspace{-0.15in}

\noindent \textbf{Geographical Region}. We evenly partition the entire urban space into $R$ geographical regions with grid-based map segmentation. Each region is our target spatial unit for predicting crime occurrence. In the urban space, both local and global region-wise dependencies exist among regions. \\\vspace{-0.15in}

\noindent \textbf{Urban Crime Data}. The data of crime occurrence reports is collected with the spatial-temporal information formatted as $\textless$crime type, timestamp, longitude, latitude$\textgreater$. Each crime report is mapped into a specific geographical region based on its coordinates. The multi-dimensional urban crime data can be formatted as a three-way tensor $\textbf{X}\in\mathbb{R}^{R\times T\times C}$, where $R$, $T$ and $C$ represents the number of regions (\ie, geographical areas) indexed $r$, time slots (\eg, days) indexed $t$ and crime types (\eg, burglary, robbery, larceny, etc) indexed $c$, respectively. Each entry $\textbf{X}_{r,t,c}$ in tensor $\textbf{X}$ represents the number of happened crime cases with type $c$ at region $r$ during the $t$-th time slot. For each region $r$, we can construct the type-specific crime sequence $\textbf{X}_{r,c} \in \mathbb{R}^{T}$. \\\vspace{-0.15in}

\noindent \textbf{Task Formalization.} Based on the above definitions, we can formalize the problem of urban crime prediction as follows:\\\vspace{-0.15in}

\noindent \textbf{Input}: the spatial-temporal crime tensor $\textbf{X}\in\mathbb{R}^{R\times T\times C}$.
\noindent \textbf{Output}: a predictive model to effective infer the number of crime cases for each region with different crime types in the future time slot $T+1$. The forecasting results can be formalized as a matrix $\hat{\textbf{X}}_{T+1}\in\mathbb{R}^{R\times C}$.

\section{Methodology}
\label{sec:solution}

\begin{figure}
    \centering
    \includegraphics[width=0.95\columnwidth]{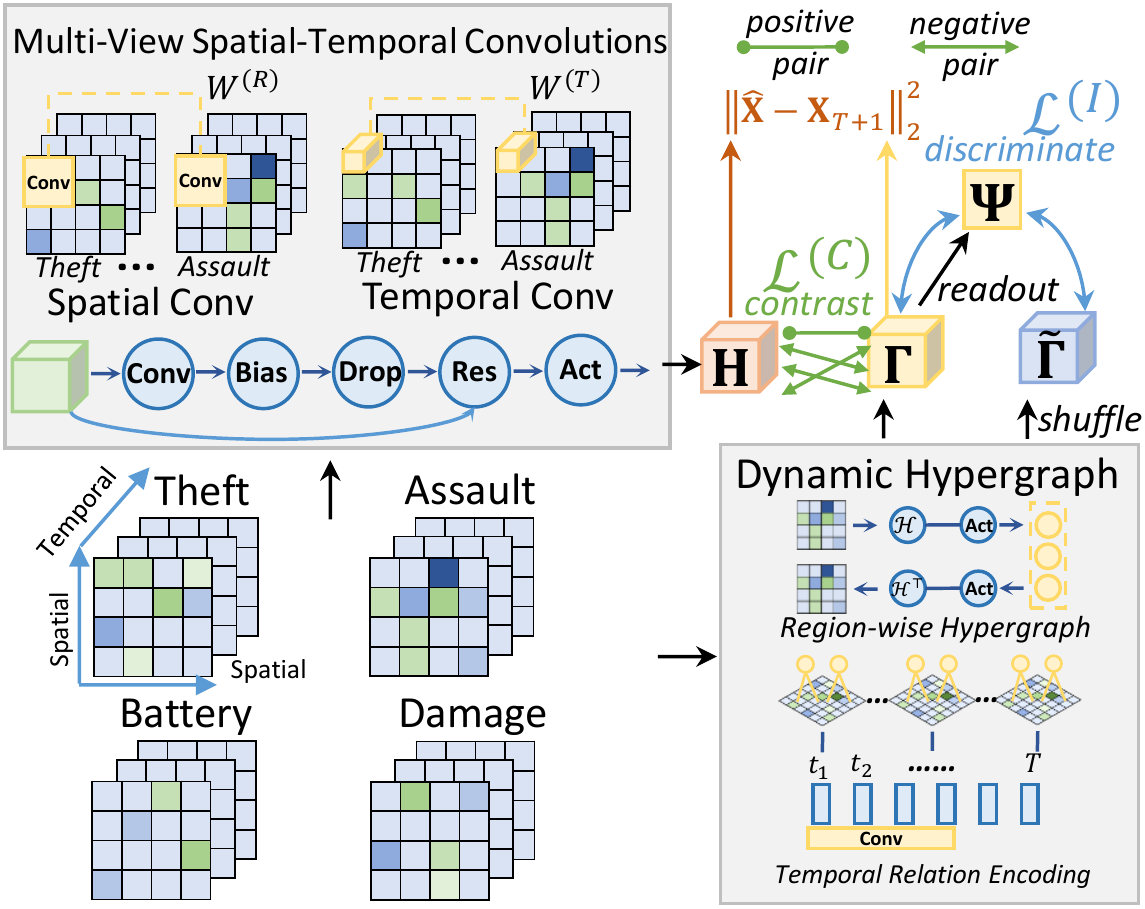}
    \caption{Overall Architecture of the proposed \model.}
    \vspace{-0.1in}
    \label{fig:fra}
\end{figure}

This section presents the technical details of our proposed \model\ crime prediction framework. There are three key learning components in our framework: (1) Multi-view spatial-temporal convolution encoder that offers initialized representations with the preservation of spatial-temporal patterns. (2) Hypergraph global dependency modeling component that refine the region embeddings by injecting high-order relationships across all regions. (3) The dual-stage self-supervised learning component that provides data augmentation with contrastive objectives from the aspect of graph-structured mutual information. The overall model architecture is illustrated in Figure~\ref{fig:fra}.\\\vspace{-0.12in}


\begin{table}[t]
    \caption{Descriptions of Key Notations}
    \label{tab:data}
    \centering
    \footnotesize
	\setlength{\tabcolsep}{0.6mm}
    \begin{tabular}{c | c}
        \hline
        Notations & Description \\
        \hline
        $R$, $T$, $C$ & Number of regions, time slots, crime types. \\
        $r$, $t$, $c$ & Index of regions, time slots, crime types.\\
        $\textbf{X}\in\mathbb{R}^{R\times T\times C}$ & Three-way urban crime tensor.\\
        $\textbf{X}_{r,t,c}$ & Number of reported crimes at region $r$, time $t$, type $c$.\\
        $\textbf{X}_{r,c} \in \mathbb{R}^{T}$ & Region- and type-specific crime sequence.\\
        $\textbf{H}_{t,c}^{(R)} \in\mathbb{R}^{d}$ & Representations with hierarchical convolutional encoder.\\
        $\textbf{H}_{r,c}^{(T)} \in\mathbb{R}^{d}$ & Representations with temporal dependency modeling.\\
        $\mathbf{\Gamma}_{t}^{(R)}\in\mathbb{R}^{RC\times d}$ & Representations with hypergraph dependency modeling.\\
        $\mathbf{\Psi}_{t,c}\in\mathbb{R}^{d}$ & Spatial graph-level representations.\\
        $\mathcal{L}^{(I)}$ & Hypergraph Infomax optimized objective.\\
        $\mathcal{L}^{(C)}$ & Cross-view contrastive optimized objective.\\
        \hline
    \end{tabular}
    \label{tab:notations}
\end{table}

\subsection{Crime Embedding Layer}
We first design an embedding layer to generate initial representation for each geographical region based on its crime occurrence distributions. In particular, we assign a randomly-initialized embedding $\textbf{e}_c\in\mathbb{R}^d$ (with dimensionality of $d$) for each crime type $c$. We generate the initial representation $\textbf{e}_{r,t,c} \in\mathbb{R}^d$ for spatial-temporal crime pattern with $c$-th type at region $r$ during the $t$-th time slot as follows:
\begin{align}
    \label{eq:initialize}
    \textbf{e}_{r,t,c} = \text{Z-Score}(\textbf{X}_{r,t,c})\cdot \textbf{e}_c = \frac{\textbf{x}_{r,t,c}-\mu}{\sigma} \cdot \textbf{e}_c
\end{align}
\noindent Here, we apply the $\text{Z-Score}$ normalization function with $\mu$ and $\sigma$ as the mean and standard variation of tensor $\textbf{X}$. 

\subsection{Multi-View Spatial-Temporal Convolution Encoder}
In our \model\ framework, we develop the multi-view spatial-temporal convolutional network to encode not only the crime dependencies among the neighboring geographical regions and consecutive time slots, as well as the latent relationships across different crime categories. \\\vspace{-0.12in}

\subsubsection{\bf Type-aware Spatial Crime Pattern Encoding}
We propose to jointly map spatial relations and type-wise crime dependence into the same latent representation space. In real-life urban scenarios, the occurrence of crimes with different types are inter-dependent in complex ways. For example, violent crimes (\eg, robbery and assault) are more likely to happen due to the deficiency of police resources at certain regions. Hence, it is of great importance to capture both the spatial crime context and implicit type-wise crime dependence. To achieve this goal, we design a hierarchical convolutional encoder which is formally presented as follows:
\begin{align}
    \label{eq:conv_spa}
    \textbf{H}_{t,c}^{(R)} = \sigma(\delta(\textbf{W}_c^{(R)} * \textbf{E}_{t} + \textbf{b}^{(R)}_c) + \textbf{E}_{t,c})
\end{align}
\noindent where $\mathbf{E}_{t,c} \in \mathbb{R}^{R \times d}$, $\mathbf{H}_{t,c}^{(R)} \in \mathbb{R}^{R \times d}$. Specifically, $\mathbf{E}_t \in \mathbb{R}^{I\times J\times C \times d}$, $I$ and $J$ represents the number of regions corresponding to the row and column dimension in the spatial region map in a city. $\textbf{W}_c^{(R)}\in\mathbb{R}^{L_{(I)} \times L_{(J)} \times C}$ and $\textbf{b}^{(R)}_c\in\mathbb{R}^{d}$ are the convolution kernel and bias terms for the $c$-th type of crimes. Here, $(R)$ indicates the spatial dimension corresponding to different regions. $L_{(I)}$ and $L_{(J)}$ denote the kernel size along the row and column dimension, respectively. $*$ denotes the convolution operation. $\delta(\cdot)$ and $\sigma(\cdot)$ represent the dropout and LeakyReLU function, respectively. In \model, we further adopt the residual connection with element-wise addition over the previous embedding $\textbf{E}_{t,c}$ to relieve gradient vanishing~\cite{he2016deep}. We stack two layers of the above designed convolutions to yield representation vectors $\textbf{H}^{(R)}\in\mathbb{R}^{R\times T\times C\times d}$ which simultaneously preserve spatial and semantic crime dependence. \\\vspace{-0.12in}

\subsubsection{\bf Temporal Crime Dependency Modeling}
To model the temporal dependency of crime occurrence across different time slots, we propose to aggregate the cross-time crime patterns with the temporal convolutional network. Formally, the aggregation process is given as below:
\begin{align}
    \label{eq:conv_tem}
    \textbf{H}_{r,c}^{(T)} = \sigma(\delta(\textbf{W}_c^{(T)} * \textbf{H}_{r}^{(R)} + \textbf{b}^{(T)}_c) + \textbf{H}_{t,c}^{(R)})
\end{align}
\noindent where $\textbf{H}_{r}^{(R)}\in\mathbb{R}^{T\times C\times d}, \textbf{H}_{t,c}^{(R)}\in\mathbb{R}^{d}$ are embeddings learned from type-aware spatial pattern encoder. $(T)$ refers to the temporal dimension. The temporal convolution kernel is defined as $\textbf{W}_c^{(T)}\in\mathbb{R}^{L_{(T)}\times C}$. The representation $\textbf{H}^{(T)} \in \mathbb{R}^{R\times T\times C\times d}$ is generated with applying two layers of our temporal convolution kernel to capture the time-evolving crime patterns.

\subsection{Hypergraph Global Dependency Modeling}
In addition to encoding local spatial relationships among nearby regions, global dependencies with respect to crime occurrence patterns also serve as key factors for accurate crime predictions. Specifically, two regions with similar urban functionality (\eg, residential area, shopping mall) can also be highly correlated even they are distant in geographical urban space. Meanwhile, the skewed crime data distribution has an adverse effect on the prediction of low-frequency crime regions. In this component, we argument our \model\ model to capture the global-level cross-region dependencies and alleviate the skewed distribution issue, by proposing a hypergraph learning framework.\\\vspace{-0.12in}

\subsubsection{\bf Region-wise Hypergraph Relation Encoding}
In our hypergraph relation learning paradigm, we propose to automatically and explicitly capture the cross-region crime dependency under a trainable hypergrpah structures. Inspired by recent advances in hypergraph neural networks~\cite{feng2019hypergraph,jiang2019dynamic}, we leverage the hyperedges as the intermediate hubs to connect a set of regions. By doing so, different regions can interact altogether with high-order connections. Here, we define the number of hyperedges as $H$. In our hypergraph learning framework, the region-wise relationships can be captured through the hypergraph-guided message passing schema between individual regions and hyperedges. In this regard, hyperedges serve as latent representation channels to preserve relation semantics from different dimensions. Formally, we define our hypergraph message passing schema with the the following form:
\begin{align}
    \label{eq:hgnn}
    \mathbf{\Gamma}_{t}^{(R)} = \sigma( \mathcal{H}_t^\top \cdot \sigma(\mathcal{H}_t \cdot \textbf{E}_{t}))
\end{align}
\noindent where $\mathcal{H}_t\in\mathbb{R}^{H\times RC}$ represents the learnable dependency structures between regions and hyperedges for information propagation over hypergraph. Note that $\mathcal{H}_t$ relates to the $t$-th time slot, by doing which \model\ is able to capture the time-evolving characteristics in global connectivity. To comprehensively consider regional features from different categorical perspectives, here we use $\textbf{E}_t\in\mathbf{R}^{RC\times d}$ which contains the embedding vectors for all region-category combinations. Here, $\sigma(\cdot)$ denotes the LeakyReLU activation function. With the design of hyperedge-based embedding propagation, \model\ allows the encoding of cross-region crime dependency with the capability of global context awareness by generating $\mathbf{\Gamma}_{t}^{(R)}\in\mathbb{R}^{RC\times d}$ under the global urban space. By effectively encoding the relevance between different regions and latent hyperedge representations, regions with similar crime distributions across the entire city can be correlated for knowledge transfer with the injecting global urban context. By doing so, the skewed distribution problem can be alleviated well.\\\vspace{-0.12in}

\subsubsection{\bf Temporal Relation Encoding}
To inject temporal context of crime patterns into our embedding paradigm, we adopt the temporal convolutional network with fusion kernel size to perform information aggregation along temporal dimension. Formally, our temporal relation encoder works as follows:
\begin{align}
    \label{eq:global_tem}
    \mathbf{\Gamma}_{r,c}^{(T)} = \sigma(\delta(\textbf{V} * \mathbf{\Gamma}_{r,c}^{(R)} + \textbf{c}))
\end{align}
\noindent where $\textbf{V}\in\mathbb{R}^{L_{(R)}'\times 1}$ and $\textbf{c}\in\mathbb{R}^{d}$ are trainable transformation parameters for temporal convolutions. $\mathbf{\Gamma}_{r,c}^{(T)}\in\mathbb{R}^{T\times d}$ contains embedding vectors for all $T$ time slots for the $r$-th region and $c$-th crime type.

\subsection{Dual-Stage Self-Supervision Learning Paradigm}
In crime prediction, one key challenge lies in the effective learning of spatial-temporal dependencies with sparse crime data. To tackle this challenge, we augment the crime pattern representation in \model\ with a dual-stage self-supervision learning paradigm, which contains two key components: (1) data augmentation with hypergraph infomax network that enhances the global context learning; (2) a contrastive learning objective that enables the interaction between the local and global spatial-temporal dependency modeling.  \\\vspace{-0.12in}

\subsubsection{\bf Hypergraph Infomax Network}
Inspired by the effectiveness of self-supervised learning for data augmentation in computer vision~\cite{chen2021pre}, nature language processing~\cite{kang2020self}, we design a hypergraph infomax network to enhance the main embedding space of crime data with auxiliary task. Motivated by the graph encoding function in~\cite{velickovic2019deep}, our hypergraph infomax network designs a hypergraph learning task with the mutual information agreement between node- and graph-level spatial-temporal representations. In particular, we generate a corrupt hypergraph structure via randomly shuffling the region indices. Here, we define $\mathbf{\Gamma}^{(R)}$ and $\tilde{\mathbf{\Gamma}}^{(R)}$ to represent the encoded local-level region embeddings (learned with propagation function in~\ref{eq:hgnn}) from the original and corrupt hypergraph structures, respectively. Then, we apply a readout function to generate the global-level representation $\mathbf{\Psi}_{t,c}\in\mathbb{R}^{d}$:
\begin{align}
    \label{eq:global_view}
    \mathbf{\Psi}_{t,c} = \sum_{r=1}^R \mathbf{\Gamma}_{r,t,c} / R
\end{align}
\noindent where $\mathbf{\Psi}_{t,c}\in\mathbb{R}^{d}$ encodes information from all regions for the $t$-th time slot and $c$-th crime type. Finally, we train our hypergraph neural network in discriminating whether the node embeddings are from the original graph and the corrupt graph, based on $\mathbf{\Psi}_{t,c}$, formally described as:
\begin{align}
    \label{eq:infomax}
    \mathcal{L}^{(I)} = &-\sum_{r=1}^R \bigm(\log(\text{sigm}(\mathbf{\Psi}_{t,c}^\top \textbf{W}^{(I)}  \mathbf{\Gamma}^{(R)}_{r,t,c}))\nonumber\\
    &+ \log(1-\text{sigm}(\mathbf{\Psi}_{t,c}^\top \textbf{W}^{(I)}  \tilde{\mathbf{\Gamma}}^{(R)}_{r,t,c}))\bigm)
\end{align}
\noindent Here, $\text{sigm}(\cdot)$ denotes the sigmoid function. Bi-linear operations with parameters $\textbf{W}^{(I)}\in\mathbb{R}^{d\times d}$ is applied. With the regularization of $\mathcal{L}^{(I)}$, the global contextual information across the entire urban space is injected into individual region embeddings $\mathbf{\Gamma}_{r,t,c}$ for augmenting self-supervision signals.\\\vspace{-0.12in}

\subsubsection{\bf Local-Global Cross-View Contrastive Learning}
With the modeling of both local and global spatial-temporal crime patterns, we further perform the cross-view contrastive learning, to enable the integration of spatial-temporal convolutional network with hypergraph global dependency encoder. The designed contrastive learning module i) allows our local and global relation encoder (two contrastive views) collaboratively supervise with each other to mitigate the sparsity issue of crime data; ii) alleviate the interference from the involved noisy information by our spatial-temporal crime pattern modeling for robust crime data representation. To be specific, we iterate all regions and categories, and pair embedding vectors from the local relation modeling and the global relation learning as positive training pairs. While embeddings of different regions from the two views are utilized as the negative pairs. Formally, the adopted InfoNCE objective function is as follows:
\begin{align}
    \label{eq:contrast}
    \mathcal{L}^{(C)} = \sum_{r=1}^R\sum_{c=1}^C \log\frac{\exp(\cos(\bar{\mathbf{\Gamma}}_{r,t,c}, \bar{\textbf{H}}_{r,t,c}, )) }{\sum_{r'} \exp(\cos(\bar{\mathbf{\Gamma}}_{r,t,c}, \bar{\textbf{H}}_{r',t,c}))}
\end{align}

\noindent where $\bar{\textbf{H}}_{r,t,c}, \bar{\mathbf{\Gamma}}_{r,t,c} \in\mathbb{R}^d$ are embedding vectors are generated by mean-pooling over the temporal dimension. Here, $\cos(\cdot)$ denotes the cosine similarity function to measure the similarity between embeddings. With the regularization of the above contrastive loss, we enhance the discrimination ability of our \model\ model in differentiating regions with respect to their crime occurrence patterns across different types and time slots.


\subsection{Model Optimization}
In this section, we discuss the learning process of our \model\ model with the joint optimized objective, by integrating the generative and contrastive self-supervised auxiliary learning tasks. Our \model\ framework make predictions on the number of cases for the future $T+1$-th time slot by conducting mean-pooling on the previous $T$ embeddings. The process is formalized as:
\begin{align}
    \label{eq:pred}
    \hat{\textbf{X}}_{r,c}=\sum_{t=1}^T\sum_{d'=1}^d  \textbf{W}_{d'}\mathbf{\Gamma}^{(T)}_{r,t,c}(d') / T
\end{align}
\noindent The weight parameters $\textbf{W}_{d}\in\mathbb{R}^{d\times d}$ is applied to latent dimensions. In the learning process of our proposed \model\ framework, we adopt the squared error loss function and integrate it with the aforementioned self-supervised loss terms $\mathcal{L}^{(I)}$ and $\mathcal{L}^{(C)}$, to generate the joint loss $\mathcal{L}$ defined as follows:
\begin{align}
    \label{eq:loss}
    \mathcal{L}=\|\hat{\textbf{X}} - \textbf{X}_{T+1}\|_2^2
    + \lambda_1\mathcal{L}^{(I)} + \lambda_2\mathcal{L}^{(C)} + \lambda_3\|\mathbf{\Theta}\|_2^2
\end{align}
\noindent where $\hat{\textbf{X}}\in\mathbb{R}^{R\times C}$ represents the predicted number of crime cases for $R$ regions and $C$ types. $\textbf{X}_{T+1}^{R\times C}$ represents the corresponding ground-truth of crime occurrence. Here, $\lambda_1, \lambda_2, \lambda_3$ are the regularization weights for balancing loss. A weight-decay regularization term is further applied. $\|\mathbf{\Theta}\|_2^2$ represents the $L_2$ norm (or Frobenius norm) for all parametric vectors and transformation matrices $\mathbf{\Theta}$. The detailed learning process of \model\ is presented in Alg~\ref{alg:learn_alg}.

\begin{algorithm}[t]
    \small
	\caption{Learning Process of \model\ Framework}
	\label{alg:learn_alg}
	\LinesNumbered
	\KwIn{Urban crime data $\textbf{X}\in\mathbb{R}^{R\times T\times C}$, maximum epoch number $E$, regularization weight $\lambda_1,\lambda_2,\lambda_3$, learning rate $\eta$}
	\KwOut{trained parameters in $\mathbf{\Theta}$}
	Initialize all parameters in $\mathbf{\Theta}$\\
    \For{$e=1$ to $E$}{
        Calculate the initial representation $\textbf{e}_{r,t,c}$ for each region, time slot and category according to Eq~\ref{eq:initialize}.\\
        Encode the spatial crime pattern with $\textbf{H}^{(R)}$ based on the type-aware spatial convolutions according to Eq~\ref{eq:conv_spa}.\\
        Encode the temporal crime pattern with $\textbf{H}^{(T)}$ based on the type-aware temporal convolutions according to Eq~\ref{eq:conv_tem}.\\
        Model global region-wise relations with $\mathbf{\Gamma}^{(R)}$ based on hypergraph neural networks according to Eq~\ref{eq:hgnn}.\\
        Generate the global representation $\mathbf{\Psi}$ for infomax learning according to Eq~\ref{eq:global_view}.\\
        Conduct information propagation on a corrupt hypergraph structure and generate $\tilde{\mathbf{\Gamma}}^{(R)}$.\\
        Inject temporal context into the global embeddings to get $\mathbf{\Gamma}^{(T)}$ according to Eq~\ref{eq:global_tem}.\\
        Make predictions $\hat{\textbf{X}}$ according to Eq~\ref{eq:pred}.\\
        Calculate the squared error loss $\|\hat{\textbf{X}}-\textbf{X}_{T+1}\|_2^2$\\
        Calculate the infomax training loss $\mathcal{L}^{(I)}$.\\
        Calculate the contrastive learning loss $\mathcal{L}^{(C)}$.\\
        
        Combine the loss terms together to get $\mathcal{L}$.\\
        
        \For{$\theta\in\mathbf{\Theta}$}{
            $\theta=\theta-\eta\cdot\partial\mathcal{L}/\partial\theta$\\
        }
    }
    \Return all parameters $\mathbf{\Theta}$\\
\end{algorithm}

\subsection{In-Depth Analysis of \model\ Framework}
In this part, we will first show the rationality of our \model\ model from the theoretical perspective. Then, we provide the analysis of the model time complexity.\\\vspace{-0.12in}


We firstly discuss the efficacy of the contrastive learning module by analyzing the generated gradients with our contrastive optimized objectives. The contrastive learning between local and global views is able to adaptively learn from the negative samples of different training difficulties. To be specific, the gradients related to a negative sample $r'$ is calculated by:
\begin{align}
    \label{eq:ssl}
    c(r') = \left(\tilde{\textbf{H}}_{r',c}^{(T)} - (\tilde{\mathbf{\Gamma}}_{r,c}^{(T)}\tilde{\textbf{H}}_{r',c}^{(T)\top}) \tilde{\mathbf{\Gamma}}_{r,c}^{(T)} \right)
\end{align}
where $\tilde{\mathbf{\Gamma}}_{r,c}^{(T)}$ and $\tilde{\textbf{H}}_{r',c}^{(T)}$ denote the normalized vectors of $\bar{\mathbf{\Gamma}}_{r',c}^{(T)}$ and $\bar{\textbf{H}}_{r,c}^{(T)}$, respectively. By inspecting the norm of gradients in Eq~\ref{eq:ssl}, we have:
\begin{align}
    \|c(r')\|_2 \propto \sqrt{1-s^2} \exp{\frac{s}{\tau}}
\end{align}
where $s=\tilde{\mathbf{\Gamma}}_{r,c}^{(T)}\tilde{\textbf{H}}_{r',c}^{(T)\top}$ represents the similarity between the anchor embedding and the negative embedding. The anchor embedding is the learned representation of the target region. Based on the self-discrimination design, we treat the encoded embeddings from the local relation modeling and the global cross-region dependency modeling as positive training pairs, and embeddings of different regions are regarded as negative samples. Large $s$ indicates that the negative sample is similar to the anchor sample, which is also known as the hard negative samples~\cite{robinson2020contrastive}. Under certain $\tau$ settings, when the similarity $s$ increases, $\sqrt{1-s^2}\exp{\frac{s}{\tau}}$ increases dramatically, and so does the norm of the gradients. In other words, our cross-view contrastive learning framework is able to adaptively assign bigger gradients to hard negative samples. This greatly promote the training efficiency of the local-global representation learning.\\\vspace{-0.12in}

\noindent \textbf{Model Complexity Analysis}. In this part, we analyze the time complexity of our new method. Specifically, for spatial-temporal dependency encoding, \model\ takes $O(R\times T\times C\times d\times(L_{(R)} + L_{(T)}))$ time complexity for local convolutional networks, and spends $O(R\times T\times C\times d\times (H+L'_{(R)})$ for global relation modeling. To calculate $\mathcal{L}^{(C)}$ for contrastive learning, $O(R^2\times C^2\times d)$ computational cost is required to calculate the denominator, which is marginally higher than the above complexity considering the empirical hyperparameter settings. Overall, our \model\ can achieve comparable model efficiency as compared to attention-based and GNN-based spatial-temporal prediction methods.

\section{Evaluation}
\label{sec:eval}

In this section, we evaluate our proposed \model\ framework with extensive experiments on real-life urban crime datasets and answer the following research questions:

\begin{itemize}[leftmargin=*]

\item \textbf{RQ1}: How does \model\ perform for accurate crime prediction as compared to various state-of-the-art baselines? \\\vspace{-0.12in}

\item \textbf{RQ2}: What are the benefits of our designed hypergraph self-supervision learning components and how they contribute to the prediction performance? \\\vspace{-0.12in}

\item \textbf{RQ3}: Does our \model\ work robustly for geographical regions with different degrees of crime data density? \\\vspace{-0.12in}

\item \textbf{RQ4}: How do different hyperparameter settings influence \model's prediction performance? \\\vspace{-0.12in}

\item \textbf{RQ5}: How does the hypergraph-guided global spatial dependency modeling benefit the model interpretation power? \\\vspace{-0.12in}

\item \textbf{RQ6}: How efficient of our \model\ forecasting framework is when competing with different baselines?

\end{itemize}

In the following subsection, we firstly describe our experimental settings and then report the evaluation results corresponding to the above research questions.

\begin{table}[t]
\centering
\footnotesize
\caption{Statistics of Experimented Urban Crime Datasets.}
\vspace{-0.1in}
\begin{tabular}{p{1.4 cm} | p{0.55cm} | p{0.55cm} | p{0.55cm} | p{0.55cm} | p{0.55cm} | p{0.55cm} | p{0.6cm} | p{0.45cm} }
\hline
\textbf{Data} & \multicolumn{4}{c|}{\textbf{NYC-Crimes}} & \multicolumn{4}{c}{\textbf{Chicago-Crimes}} \\
\hline
Time Span & \multicolumn{4}{c|}{Jan, 2014 to Dec, 2015} & \multicolumn{4}{c}{Jan, 2016 to Dec, 2017} \\
\hline
Category & \multicolumn{2}{c|}{Burglary} & \multicolumn{2}{c|}{Robbery} & \multicolumn{2}{c|}{Theft} & \multicolumn{2}{c}{Battery} \\
\hline
Cases \# & \multicolumn{2}{c|}{31,799} & \multicolumn{2}{c|}{33,453} & \multicolumn{2}{c|}{124,630} & \multicolumn{2}{c}{99,389}\\
\hline
Category & \multicolumn{2}{c|}{Assault} & \multicolumn{2}{c|}{Larceny} & \multicolumn{2}{c|}{Damage} & \multicolumn{2}{c}{Assault}  \\
\hline
Cases \# & \multicolumn{2}{c|}{40,429} & \multicolumn{2}{c|}{85,899} & \multicolumn{2}{c|}{59,886} & \multicolumn{2}{c}{37,972} \\
\hline
\end{tabular}
\label{tab:data}
\vspace{-0.1in}
\end{table}

\subsection{Experimental Setting}

\subsubsection{\bf Dataset Description}
Our experiments are performed on two collected crime datasets from New York City (NYC) and Chicago to contain different types of crime occurrence at different locations in a city, such as Robbery, Larceny for NYC crimes; and Damage, Assault for Chicago crimes. In our experiments, we apply the $3km\times 3km$ spatial gird unit to NYC and Chicago and generate 256 and 168 disjoint spatial regions, respectively. The target resolution of predicted time period is set as day. The training and testing set are constructed with the ratio of 7:1 along with the time dimension. We tune the parameters on the validation set generated from the last 30 days of the training set. The final reported model performance is averaged over all days in the test period for all compared methods. Table~\ref{tab:data} summarizes the data statistics.\\\vspace{-0.12in}

\subsubsection{\bf Evaluation Metrics} To evaluate the accuracy of forecasting urban crimes, we utilize two metrics: \emph{Mean Absolute Error (MAE)} and \emph{Mean Absolute Percentage Error (MAPE)} which have been widely used in predicting continuous spatial-temporal data (\eg, traffic volume~\cite{liang2021fine,lin2019deepstn+} and air quality~\cite{yi2018deep}). For fair comparison and alleviating the evaluation bias, the reported prediction performance of all compared methods are averaged over all days in the test time period. Note that lower MAE and MAPE score indicates better crime prediction performance.\\\vspace{-0.15in}

\begin{table*}[t]
	\centering
	\caption{Overall performance of urban crime prediction on NYC and CHI dataset in terms of \textit{MAE} and \textit{MAPE}}
	\vspace{-0.05in}
	\label{tab:overall_performance}
    \footnotesize
	\setlength{\tabcolsep}{0.8mm}
	\begin{tabular}{|c|c|c|c|c|c|c|c|c|c|c|c|c|c|c|c|c|}
		\hline
		\multirow{3}{*}{Model} & \multicolumn{8}{c|}{New York City} & \multicolumn{8}{c|}{Chicago}\\
		\cline{2-17}
		& \multicolumn{2}{c|}{Burglary} & \multicolumn{2}{c|}{Larceny} & \multicolumn{2}{c|}{Robbery} & \multicolumn{2}{c|}{Assault} & \multicolumn{2}{c|}{Theft} & \multicolumn{2}{c|}{Battery} & \multicolumn{2}{c|}{Assault} & \multicolumn{2}{c|}{Damage}\\
		\cline{2-17}
		& MAE & MAPE & MAE & MAPE & MAE & MAPE & MAE & MAPE & MAE & MAPE & MAE & MAPE & MAE & MAPE & MAE & MAPE\\
		\hline
		\hline
		ARIMA & 0.8999 & 0.6305 & 1.3015 & 0.6268 & 0.9558 & 0.5969 & 0.9983 & 0.6198 & 1.5965 & 0.5720 & 1.3212 & 0.5792 & 0.8691 & 0.6044 & 1.0430 & 0.6134 \\
		\hline
		SVM & 1.1604 & 0.7653 & 1.4979 & 0.6417 & 1.1278 & 0.6733 & 1.1928 & 0.6964 & 1.7711 & 0.5629 & 1.3493 & 0.6027 & 1.0879 & 0.6560 & 1.1313 & 0.5721\\
		\hline
 		ST-ResNet & 0.8680 & 0.5603 & 1.1082 & 0.5329 & 0.8717 & 0.5209 & 0.9645 & 0.5749 & 1.3931 & 0.5488 & 1.1519 & 0.5719 & 0.7679 & 0.4633 & 0.9064 & 0.5018\\
        \hline
        DCRNN & 0.8176 & 0.5324 & 1.0732 & 0.5492 & 0.9189 & 0.5532 & 0.9692 & 0.5955 & 1.3699 & 0.5770 & 1.1583 & 0.5528 & 0.7639 & 0.4600 & 0.8764 & 0.4756\\
        \hline
        STGCN & 0.8366 & 0.5404 & 1.0629 & 0.5295 & 0.9035 & 0.5441 & 0.9375 & 0.5757 & 1.3628 & 0.5359 & 1.1512 & 0.5761 & 0.7963 & 0.4810 & 0.9068 & 0.4959\\
        \hline
        GWN & 0.7993 & 0.5235 & 1.0493 & 0.5405 & 0.8681 & 0.5351 & 0.8866 & 0.5646 & 1.3211 & 0.5502 & 1.1331 & 0.5503 & 0.7493 & 0.4580 & 0.8584 & 0.4850\\
        \hline
        STtrans & 0.8617 & 0.5592 & 1.0896 & 0.5478 & 0.8839 & 0.5651 & 0.9363 & 0.5679 & 1.3404 & 0.5356 & 1.1466 & 0.5684 & 0.7671 & 0.4499 & 0.8987 & 0.4842\\
        \hline
        DeepCrime & 0.8227 & 0.5508 & 1.0618 & 0.5351 & 0.8841 & 0.5537 & 0.9222 & 0.5677 & 1.3391 & 0.5430 & 1.1290 & 0.5389 & 0.7737 & 0.4616 & 0.9096 & 0.4960\\
        \hline
        STDN & 0.8831 & 0.5768 & 1.1442 & 0.5889 & 0.9230 & 0.5649 & 0.9498 & 0.5661 & 1.5303 & 0.6287 & 1.2076 & 0.5791 & 0.8052 & 0.4820 & 0.9169 & 0.4869\\
        \hline
        ST-MetaNet & 0.8285 & 0.5369 & 1.0697 & 0.5627 & 0.9214 & 0.5766 & 0.9323 & 0.5702 & 1.3369 & 0.5369 & 1.1762 & 0.5748 & 0.7904 & 0.4753 & 0.8907 & 0.4756\\
        \hline
        GMAN & 0.8652 & 0.5633 & 1.0503 & 0.5340 & 0.9234 & 0.5671 & 0.9338 & 0.5803 & 1.3235 & 0.5307 & 1.1442 & 0.5560 & 0.7852 & 0.4714 & 0.8823 & 0.4838\\
        \hline
        AGCRN & 0.8260 & 0.5397 & 1.0499 & 0.5404 & 0.9013 & 0.5383 & 0.9063 & 0.5519 & 1.3281 & 0.5304 & 1.1432 & 0.5697 & 0.7669 & 0.4612 & 0.8712 & 0.4859\\
        \hline
        MTGNN & 0.8329 & 0.5439 & 1.0473 & 0.5330 & 0.8759 & 0.5457 & 0.9090 & 0.5714 & 1.3054 & 0.5378 & 1.1307 & 0.5597 & 0.7571 & 0.4572 & 0.8667 & 0.4859\\
        \hline
        STSHN & 0.8012 & 0.5198 & 1.0431 & 0.5291 & 0.8717 & 0.5362 & 0.9169 & 0.5682 & 1.3231 & 0.5310 & 1.1348 & 0.5544 & 0.7758 & 0.4574 & 0.8741 & 0.4747\\
        \hline
        DMSTGCN & 0.8376 & 0.5485 & 1.0410 & 0.5464 & 0.8597 & 0.5403 & 0.9036 & 0.5601 & 1.3292 & 0.5291 & 1.1297 & 0.5552 & 0.8058 & 0.4759 & 0.8698 & 0.4877\\
        \hline
        \hline  
        \emph{\model} & \textbf{0.7329} & \textbf{0.4788} & \textbf{1.0316} & \textbf{0.5040} & \textbf{0.7912} & \textbf{0.4595} & \textbf{0.8484} & \textbf{0.5029} & \textbf{1.2952} & \textbf{0.4929} & \textbf{1.1016} & \textbf{0.5231} & \textbf{0.6665} & \textbf{0.3996} & \textbf{0.8446} & \textbf{0.4644}\\
        \hline
	\end{tabular}
	\vspace{-0.1in}
\end{table*}

\subsubsection{\bf Baselines for Comparison} To comprehensively evaluate our method, we compare \model\ with 15 baselines based on various spatial-temporal predictive solutions:\\\vspace{-0.12in}

\noindent \textbf{Conventional Time Series Prediction Methods}. We consider the traditional time series prediction approaches as baselines.

\begin{itemize}[leftmargin=*]

\item \textbf{ARIMA}~\cite{icdm12}: The autoregressive integrated moving average model aims to capture the correlations between observations lagged dependent variables for series prediction.\\\vspace{-0.15in}

\item \textbf{SVM}~\cite{chang2011libsvm}: The support vector machine has been used to predict periodic patterns of time series data, with the consideration of non-linear and non-stationary time series pattern.\\\vspace{-0.15in}

\end{itemize}

\noindent \textbf{Spatial-Temporal Prediction with CNNs}. Convolutional neural networks have been utilized to fuse spatial features over grid-based regions for forecasting crowd flow.

\begin{itemize}[leftmargin=*]

\item \textbf{ST-ResNet}~\cite{zhang2017deep}: This method designs convolutional network to model region-wise correlations with residual connections. Three types of time properties (\ie, closeness, period, trend) are considered for temporal modeling.\\\vspace{-0.15in}

\end{itemize}

\noindent \textbf{Spatial-Temporal Prediction via Graph Neural Networks}. Graph neural networks have become the state-of-the-art spatial-temporal prediction models which capture the spatial dependency with message passing among regions.
\begin{itemize}[leftmargin=*]

\item \textbf{DCRNN}~\cite{li2017diffusion}: This method integrates the sequence-to-sequence learning framework with the diffusional convolutional operation, to simulate the temporal and spatial dynamics with diffusion process for making prediction.\\\vspace{-0.15in}

\item \textbf{STGCN}~\cite{yubingspatio}: This method integrates the spatial graph convolutional network with the temporal gated convolutional network to generate spatial-temporal representations over graph-structured time series. Several convolutional blocks are combined in this model with the kernel size of 3. \\\vspace{-0.15in}

\item \textbf{GWN}~\cite{wu2019graph}:This method incorporates the adaptive adjacency matrix into graph convolution with 1-D dilated casual convolution to capture spatial and temporal correlations. \\\vspace{-0.15in}

\item \textbf{GMAN}~\cite{zheng2020gman}: It is a graph-based multi-attention model to encode the spatial-temporal correlations and perform the transformation between the encoder and decoder to make predictions on spatial-temporal data.\\\vspace{-0.15in}


\item \textbf{ACGRN}~\cite{bai2020adaptive}: This method uses recurrent neural network to encode temporal representations. Additionally, relations between regions are modeled with the graph convolutional network with adaptive learning methods. \\\vspace{-0.15in}


\item \textbf{MTGNN}~\cite{wu2020connecting}: This method aims to capture the spatial correlations based on a new graph neural framework without explicit graph structural information. \\\vspace{-0.15in}

\item \textbf{DMSTGCN}~\cite{han2021dynamic}: This approach enhances the graph convolutional network with dynamic and multi-faceted spatial and temporal information. In DMSTGCN, the time-aware graph constructor is applied to capture the periodicity and dependencies among road segments. \\\vspace{-0.15in}



\end{itemize}

\noindent \textbf{Hybrid Spatial-Temporal Prediction Models}. Another line of spatial-temporal prediction designs the hybrid models to encode relationships among regions and time periods.
\begin{itemize}[leftmargin=*]

\item \textbf{ST-MetaNet}~\cite{pan2019urban}: This model is a meta-learning approach which is built on GNN-based sequence-to-sequence paradigm to extract region-specific meta knowledge, and capture diverse spatial correlations.\\\vspace{-0.15in}

\item \textbf{STDN}~\cite{yao2019revisiting}: In this framework, a flow gating scheme is introduced to capture time-aware dependence between regions, a periodic shifted attention is proposed to learn temporal patterns among different time periods.

\end{itemize}

\noindent \textbf{Attention-based Crime Prediction Method}. Existing deep learning crime prediction methods are built on the attention mechanism to aggregate information from spatial and temporal dimensions. These work has validated the performance superiority of neural network-based models over conventional feature-based approaches, such as logistic regression.

\begin{itemize}[leftmargin=*]

\item \textbf{DeepCrime}~\cite{huang2018deepcrime}: It is a representative crime prediction baseline which first uses the recurrent neural network to encode temporal embeddings of crime occurrences across time. Then, attention mechanism is utilized to further aggregate temporal representations with the attentional weights.\\\vspace{-0.15in}

\item \textbf{STtrans}~\cite{2020hierarchically}: It explores the sparse crimes by stacking two layers of Transformer to encode spatial-temporal relationships across locations and time. Self-attention with query/key transformations is adopted for spatial and temporal information aggregation.

\end{itemize}

\noindent \textbf{Graph-based Crime Prediction Model}. We also compare our \model\ with the recently developed crime prediction framework under a graph learning architecture.

\begin{itemize}[leftmargin=*]

\item \textbf{STSHN}~\cite{xiaspatial}. This method performs the spatial message passing among different geographical regions based on the hypergraph connections between regions. The region hypergraph is constructed in a stationary manner. The number of spatial path aggregation layers is set as 2. We set the number of hypergraph channels as 128 to be consistent with our \model\ model for fair comparison. \\\vspace{-0.12in}

\end{itemize}

\subsubsection{\bf Hyperparameter Settings}
Our \model\ is optimized with Adam optimizer with the learning rate of 0.001. We search the hidden state dimensionality from the range \{$2^2$, $2^3$, $2^4$, $2^5$\}. For our multi-view spatial-temporal convolutions, the kernel size is set as 3 with two convolutional layers. For our hypergraph dependency learning component, we configure \model\ with 128 hyperedges for cross-region embedding propagation. We stack 4 convolutional layers for long-term temporal context modeling. The batch size is selected from $\{4, 8, 16, 32\}$. The weight for regularization terms $\lambda_1, \lambda_2, \lambda_3$ are selected from the range of $(0.0, 1.0)$. 


\subsection{Performance Comparison (RQ1)}
\begin{figure*}[t]
    \centering
    \includegraphics[width=1.9\columnwidth]{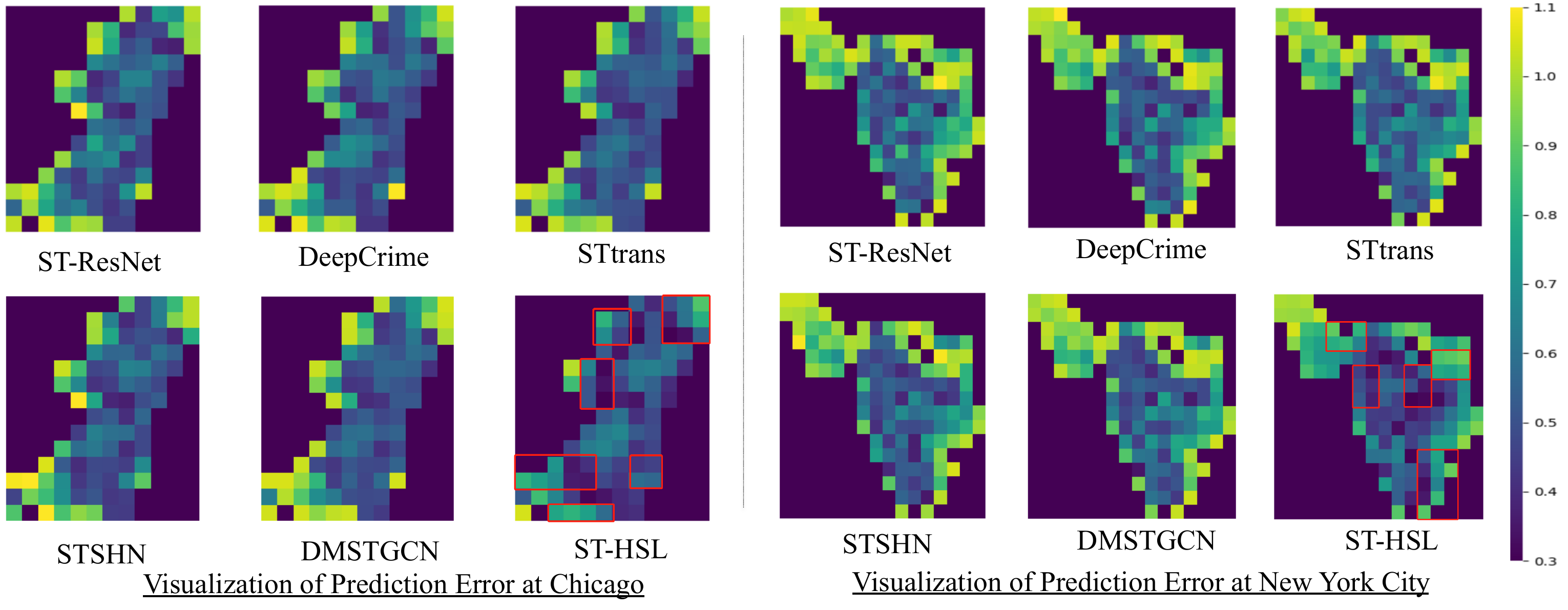}
    \vspace{-0.1in}
    \caption{Prediction error Visualization of different methods over geographical regions in the entire urban space of NYC and Chicago. Best view in color.}
    \label{fig:visuallization_error}
\end{figure*}

Table~\ref{tab:overall_performance} shows the performance comparison results between our \model\ method and different types of baselines for urban crime prediction. We summarize our findings as:


\begin{itemize}[leftmargin=*]

\item \model\ outperforms various spatial-temporal prediction methods in all cases, which suggests the superiority of our hypergraph contrastive learning paradigm in supplementing the spatial-temporal dependency modeling for crime predictions, with dual-stage self-supervised learning. Specifically, we attribute these significant improvements to: i) Benefiting from our hypergraph dependency encoder, \model\ is able to capture the holistic crime patterns and preserve global spatial-temporal signals across the entire urban space; ii) The designed dual-stage self-supervised learning paradigm incorporates auxiliary self-supervision signals, which can provide informative spatial-temporal representations with sparse crime data. \\\vspace{-0.12in}

\item While GNN-based methods (DCRNN, STGCN, GMAN) capture spatial dependence with the high-order information propagation over the region graph, they approach the crime prediction task under the supervised learning architecture. However, the sparse supervision labeled data limit their discrimination ability to produce quality spatial-temporal representations under the highly sparse and skewed distributed crime data. The performance gap between \model\ and attention-based crime prediction models (DeepCrime, STtrans), suggests the rationality of our global context enhancement for modeling spatial relationships with respect to crime occurrences. Additionally, by comparing our \model\ framework with the baseline STSHN, the performance improvement indicates that the augmented hypergraph learning tasks are critical for enhancing the encoding of complex crime patterns with effective self-supervised regularization. \\\vspace{-0.12in}

\item To have a better understanding of our prediction results, we visualize the crime prediction results in terms of MAPE over the entire urban spatial space for both New York City and Chicago. The visualization results are presented in Figure~\ref{fig:visuallization_error}. We can observe that \model\ can provide obvious better prediction performance at the highlighted spatial areas. Another observation is that our \model\ still achieves significant superior performance compared with state-of-the-art baselines at regions with relatively fewer crime occurrence. \\\vspace{-0.12in}

\item With the joint performance analysis across different crime types, we notice that the improvement on sparse crime types (\eg, Burglary, Robbery, Assault) is more significant than that on relatively dense types (\eg, Larceny, Battery) of urban crimes. Most baselines are easily biased and relatively unstable than our \model\ for predicting different categories of crime data. These observations further validate the effectiveness of incorporating auxiliary self-supervision signals from both generative and contrastive views under the hypergraph-guided learning paradigm, to guide the sparse crime pattern representation with effective augmentations.


\end{itemize}

\begin{figure}
    \centering
    \subfigure[Evaluation Results on NYC Data]{
        \centering
        \includegraphics[width=0.45\columnwidth]{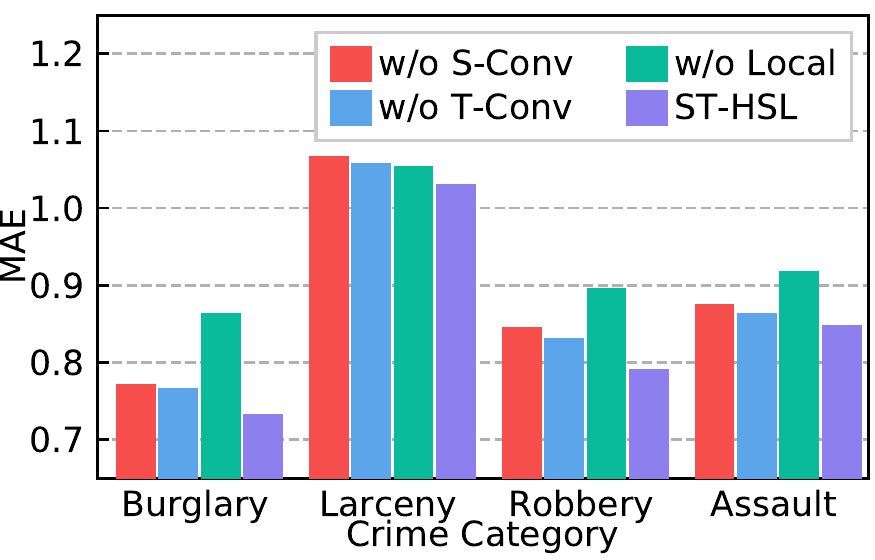}
        \includegraphics[width=0.45\columnwidth]{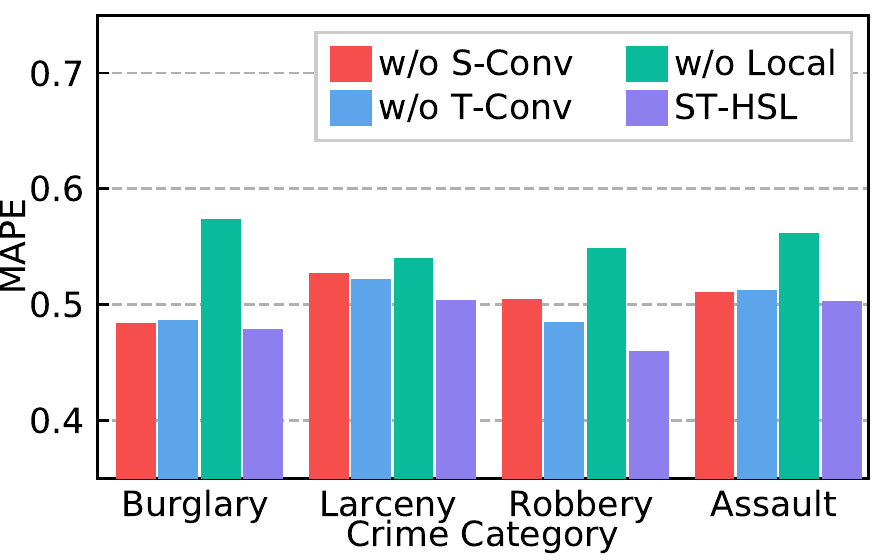}
    }
    \subfigure[Evaluation Results on Chicago Data]{
        \centering
        \includegraphics[width=0.45\columnwidth]{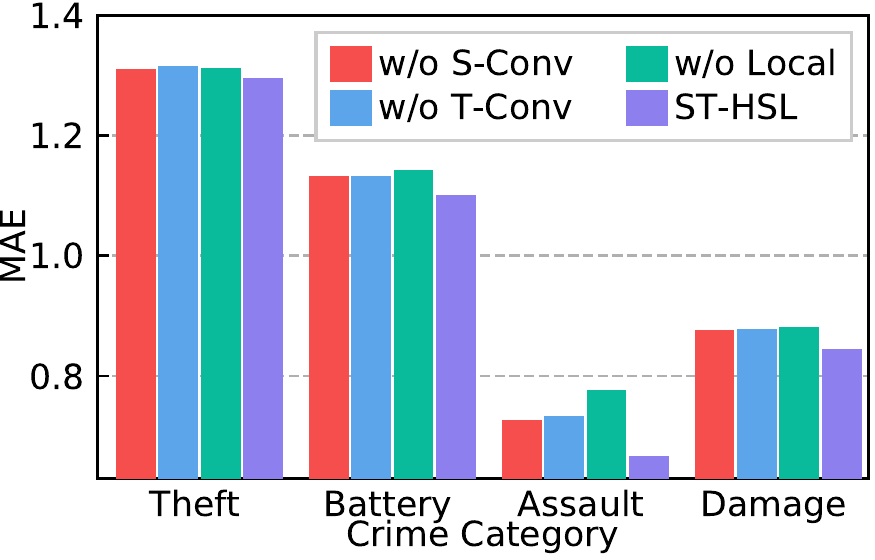}
        \includegraphics[width=0.45\columnwidth]{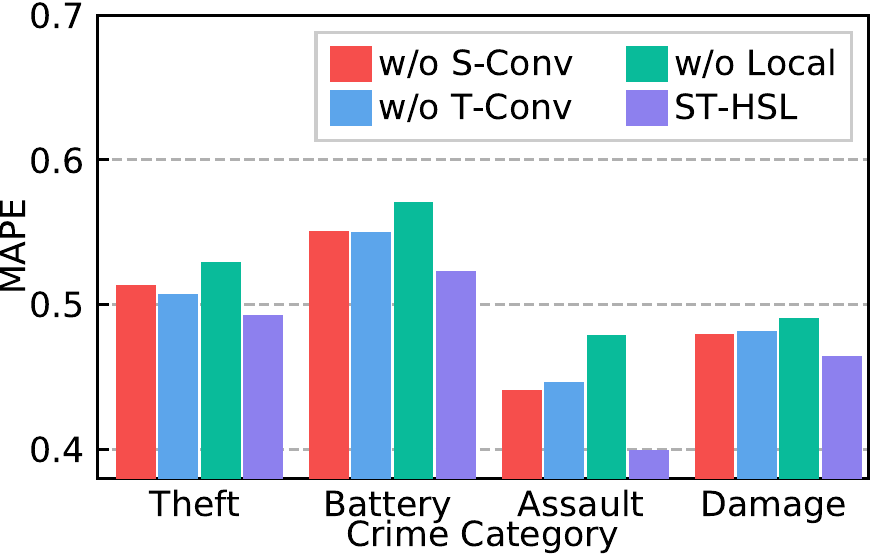}
    }
    \vspace{-0.05in}
    \caption{Module ablation study on multi-view spatial-temporal relation encoder in our \model\ framework, in terms of \textit{MAE} and \textit{MAPE}.}
    \label{fig:ablation_ssl}
    \vspace{-0.15in}
\end{figure}

\subsection{Model Ablation and Effectiveness Analyses (RQ2)}

We further explore how different components in \model\ contribute to the prediction performance of \model. In particular, the model ablation study is conducted to investigate the benefits of \model's key components, \ie, multi-view spatial-temporal encoder, dual-stage self-supervised learning paradigm. The results are reported in Figure~\ref{fig:ablation_ssl} and Table~\ref{tab:ablation_SSL}. \\\vspace{-0.1in}

\subsubsection{\bf Multi-View Spatial-Temporal Convolution} To verify the effectiveness of our spatial-temporal convolution network for modeling the multi-view dependencies, we generate three variants: ``w/o S-Conv'', ``w/o T-Conv'', ``w/o C-Conv'' by disabling the local relation representation for spatial, temporal, and category, respectively. Furthermore, we remove entire multi-view local encoder to generate the variant termed as ``w/o Local''. According to results in Figure~\ref{fig:ablation_ssl}, by comparing with the aforementioned model variants, \model\ demonstrates its effectiveness to distill useful knowledge from different views corresponding to spatial, temporal, and semantic information of crime data. Each view-specific encoded semantic is complementary with each other, which has positive effect on the overall crime prediction performance.\\\vspace{-0.1in}

\subsubsection{\bf Dual-Stage Self-Supervised Learning} 
We also perform ablation study to evaluate the effectiveness of our dual-stage self-supervised learning architecture, so as to enhance the spatial-temporal representation based on the self-discrimination of regions. In this part, we generate seven contrast method variants: 1) ``w/o Hyper''. We leave the region-wise hypergraph relation unexplored and rely on the local spatial encoder to make prediction; 2) ``w/o GlobalTem''. We leave the global temporal encoder unexplored and rely on the local temporal encoder to make prediction; 3) ``w/o Infomax''. We disable the hypergraph infomax network for global context injection with auxiliary self-supervision signals. 4) ``w/o ConL''. This variant does not include the cross-view contrastive learning to enable the interaction between the local and global spatial dependency encoder in our framework. 5) ``w/o Global''. This variant set same as 4) without contrastive learning. Differently, we only use local encoder to make prediction. 6) ``Fusion w/o ConL''. We use a fusion layer to aggregate the the local and global view embeddings to make prediction, without the incorporation of cross-view contrastive learning paradigm. 


As we can see in Table~\ref{tab:ablation_SSL}, with the incorporation of our self-supervised learning paradigm, \model\ performs the best in most evaluation cases. This again emphasizes the benefits of exploring self-supervision signals with hypergraph-guided relational learning augmentation, to alleviate the crime data sparsity issue and skewed distribution issue for better spatial-temporal representations. In our joint learning framework--\model, our hypergraph information network enhances the region-wise dependency modeling, by integrating a self-supervised hypergraph learning task into the embedding space of crime patterns. In the comparative experiment with the fusion layer, we can find that the effect of aggregating local and global view information by using the contrastive learning method is better than using the fusion layer. We attribute such performance improvement to two aspects: i) Contrastive learning allows the model to explore useful information from the data itself, which is helpful to generate more robust feature representation. ii) Both the fusion layer and cross-view contrastive learning can establish the aggregation and balance between local and global features. However, only using a supervised loss to guide the fusion layer to make trade-offs may be no easy. By adding additional infoNCE loss, a guide information can be generated intuitively, which can better aggregate local and global features. Moreover, the designed hypergraph contrastive learning mechanism with effective augmentation that incorporates both local and global semantic for robust spatial-temporal representation. Through our hypergraph self-supervised learning paradigm, the main and augmented representation tasks are mutually enhanced each other to produce better region embeddings.


\begin{table}[t]
    \centering
    \caption{Module ablation study on the hypergraph dual-stage self-supervised learning paradigm.}
    \vspace{-0.05in}
    \label{tab:ablation_SSL}
    \begin{tabular}{l|c|c|c|c}
        \hline
        \multirow{2}{*}{Model} & \multicolumn{4}{c}{NYC-Data} \\
        \cline{2-5}
        & \multicolumn{1}{c|}{Burglary} & \multicolumn{1}{c|}{Larceny} & \multicolumn{1}{c|}{Robbery} & \multicolumn{1}{c}{Assault} \\
        \cline{2-5}
        \hline
        w/o Hyper & 0.7929 & 1.0380 & 0.8567 & 0.9010 \\
        w/o GlobalTem & 0.8531 & 1.0866 & 0.9226 & 0.9285 \\
        w/o Infomax & 0.7512 & 1.0382 & 0.8338 & 0.8603 \\
        w/o ConL & 0.8938 & 1.0757 & 0.9345 & 0.9529 \\
        w/o Global & 0.7876 & 1.0583 & 0.8740 & 0.9472 \\
        Fusion w/o ConL & 0.7939 & 1.0438 & 0.8551 & 0.8877 \\
        \hline
        \emph{\model} & \textbf{0.7329} & \textbf{1.0316} & \textbf{0.7912} & \textbf{0.8484} \\
        \hline
        \multirow{2}{*}{Model} & \multicolumn{4}{c}{Chicago-Data} \\
        \cline{2-5}
        & \multicolumn{1}{c|}{Theft} & \multicolumn{1}{c|}{Battery} & \multicolumn{1}{c|}{Assault} & \multicolumn{1}{c}{Damage}\\
        \cline{1-5}
        w/o Hyper & 1.3041 & 1.1214 & 0.7134 & 0.8657 \\
        w/o GlobalTem & 1.3147 & 1.1703 & 0.7208 & 0.8699 \\
        w/o Infomax & 1.2972 & 1.1196 & 0.7000 & 0.8507 \\
        w/o ConL & 1.3211 & 1.1598 & 0.7694 & 0.8849 \\
        w/o Global & 1.3053 & 1.1351 & 0.7318 & 0.8626 \\
        Fusion w/o ConL & 1.3010 & 1.1365 & 0.7482 & 0.8592 \\
        \hline
        \emph{\model} & \textbf{1.2952} & \textbf{1.1016} & \textbf{0.6665} & \textbf{0.8446} \\
        \hline
    \end{tabular}
    \vspace{-0.05in}
\end{table}

\begin{figure*}
    \centering
    \includegraphics[width=0.5\textwidth]{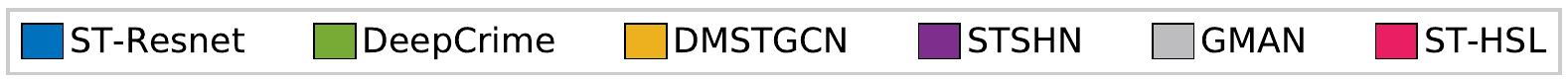}
    
    \subfigure[Evaluation Results on NYC Data for Burglary, Larceny, Robbery and Assault]{
        \centering
        \includegraphics[width=0.115\textwidth]{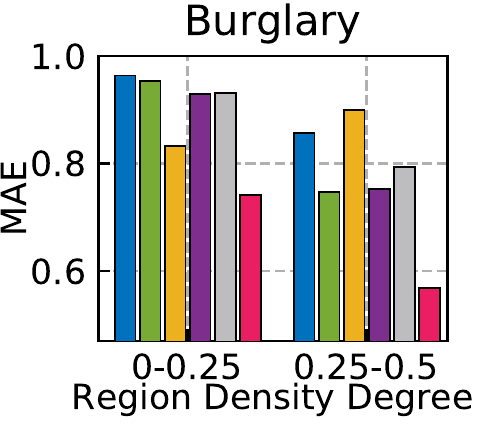}
        \includegraphics[width=0.115\textwidth]{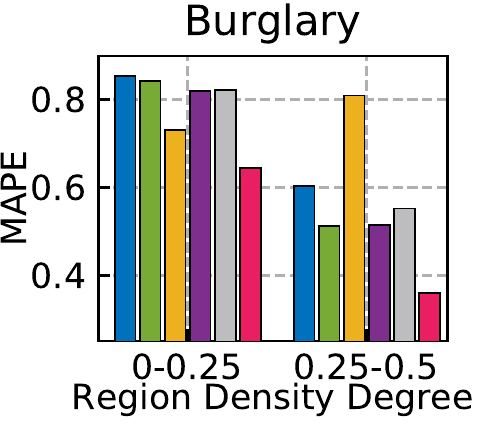}
        \includegraphics[width=0.115\textwidth]{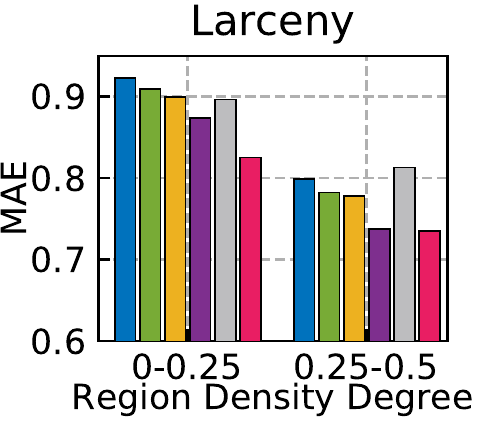}
        \includegraphics[width=0.115\textwidth]{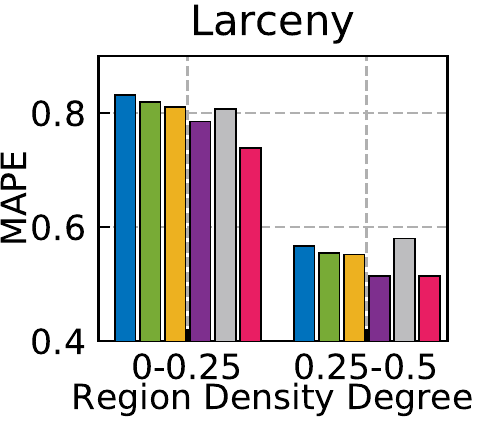}
        \includegraphics[width=0.115\textwidth]{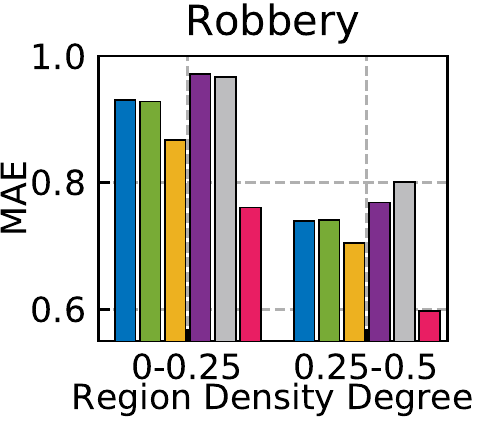}
        \includegraphics[width=0.115\textwidth]{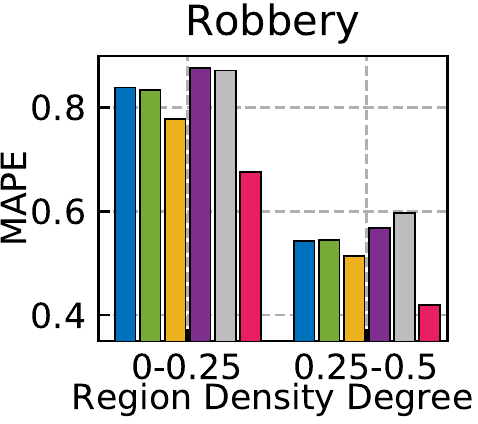}
        \includegraphics[width=0.115\textwidth]{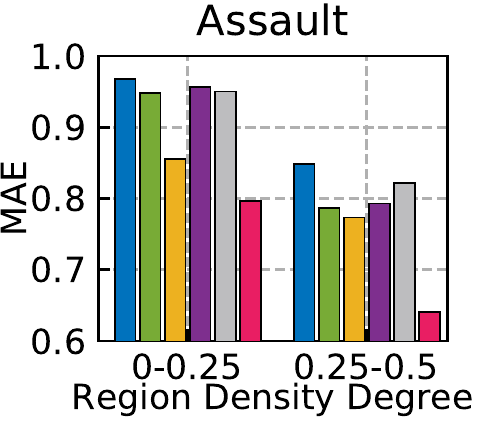}
        \includegraphics[width=0.115\textwidth]{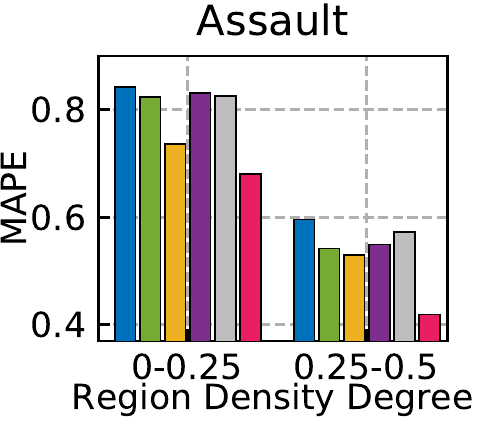}
    }
    \subfigure[Evaluation Results on Chicago Data for Theft, Battery, Assault and Damage]{
        \centering
        \includegraphics[width=0.115\textwidth]{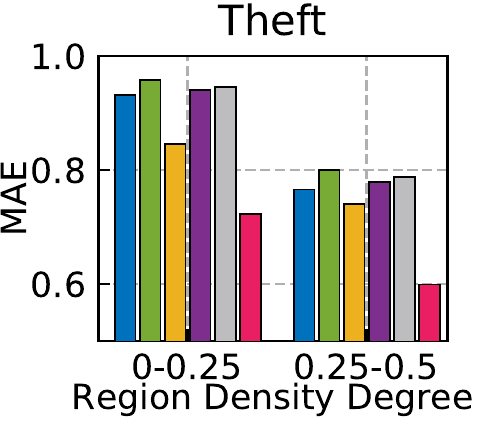}
        \includegraphics[width=0.115\textwidth]{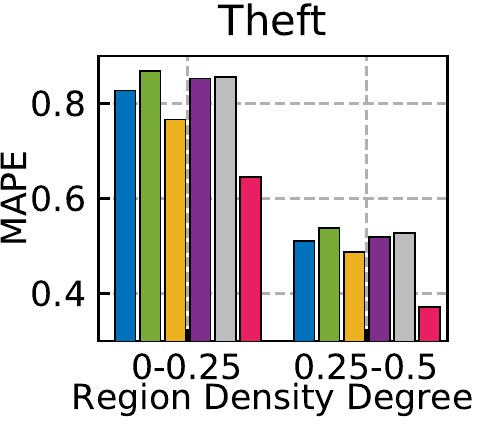}
        \includegraphics[width=0.115\textwidth]{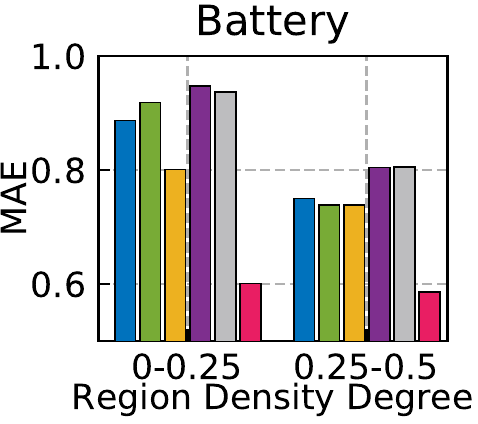}
        \includegraphics[width=0.115\textwidth]{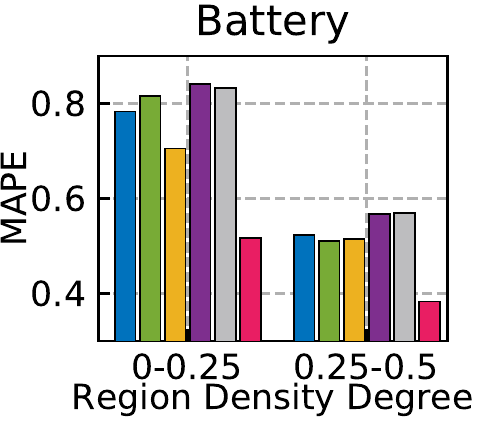}
        \includegraphics[width=0.115\textwidth]{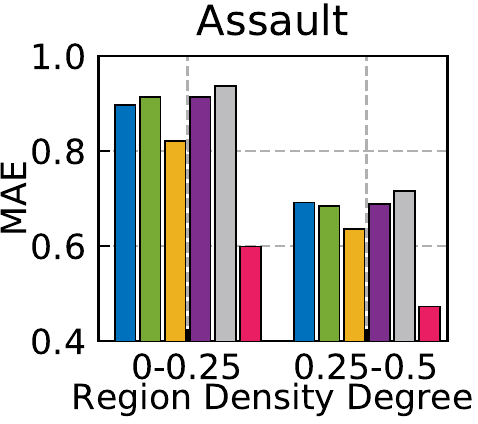}
        \includegraphics[width=0.115\textwidth]{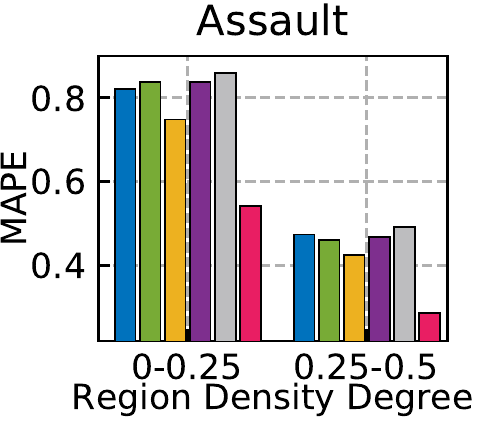}
        \includegraphics[width=0.115\textwidth]{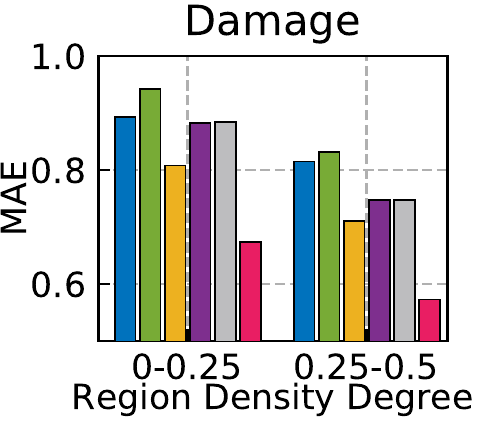}
        \includegraphics[width=0.115\textwidth]{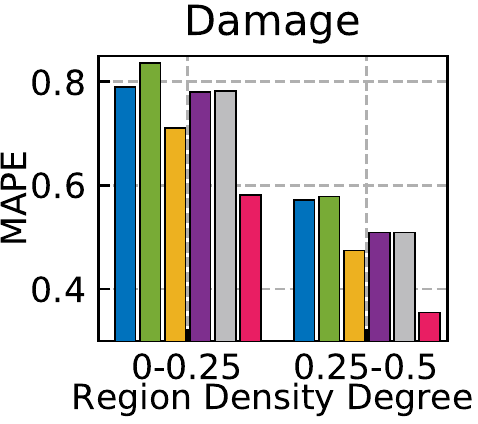}
    }
    
    \caption{Robustness study of our \model\ to data sparsity. X-axis represents the region crime density degree, \ie, the ratio of non-zero elements in region-specific crime sequence. Y-axis represents the prediction performance (measured by MAE and MAPE) of different compared methods.}
    \label{fig:sparsity}
\end{figure*}

\begin{figure*}
    \centering
    \includegraphics[width=0.175\textwidth]{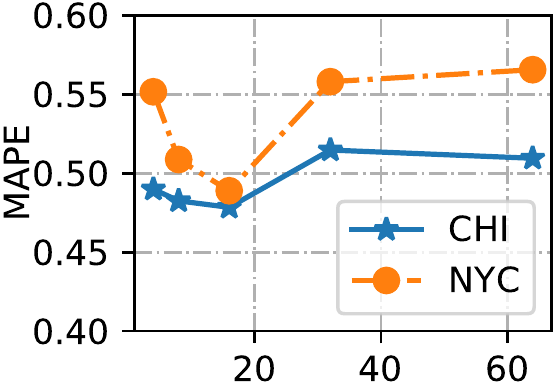}\quad
    \includegraphics[width=0.175\textwidth]{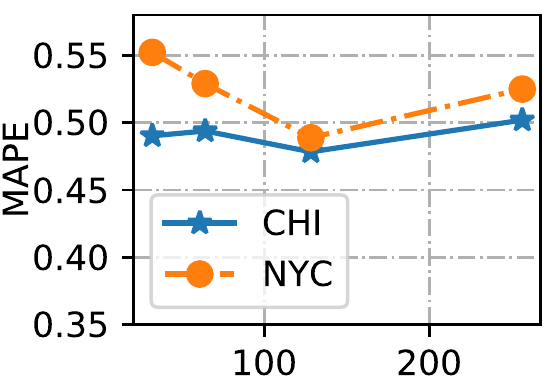}\quad
    \includegraphics[width=0.175\textwidth]{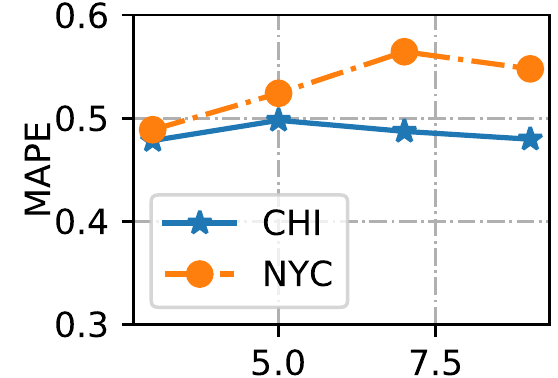}\quad
    \includegraphics[width=0.175\textwidth]{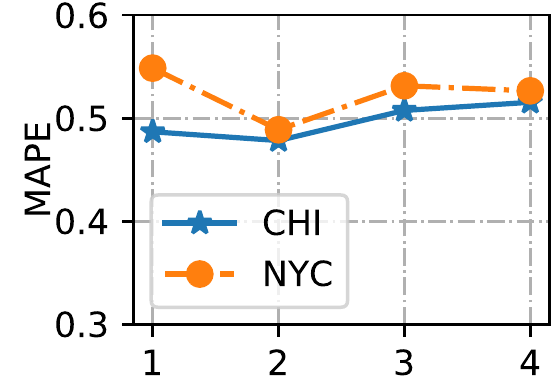}\quad
    \includegraphics[width=0.175\textwidth]{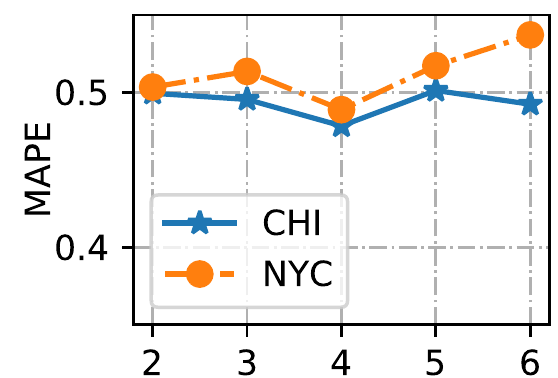}\quad
    \subfigure[\# Hidden Units]{
        \centering
        \includegraphics[width=0.175\textwidth]{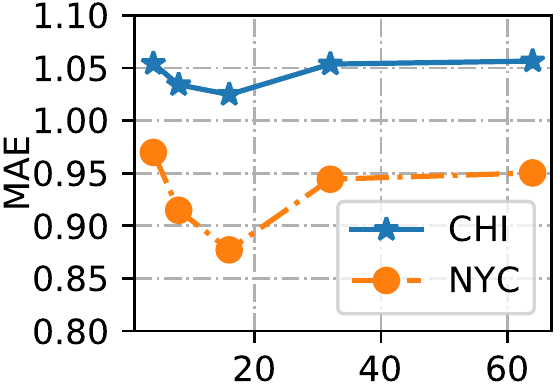}
    }
    \subfigure[\# Hyperedges]{
        \centering
        \includegraphics[width=0.175\textwidth]{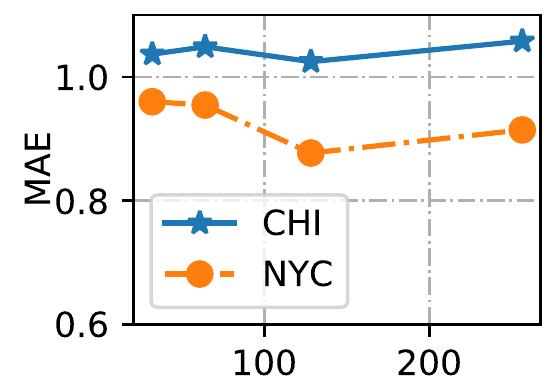}
    }
    \subfigure[Kernel Size]{
        \centering
        \includegraphics[width=0.175\textwidth]{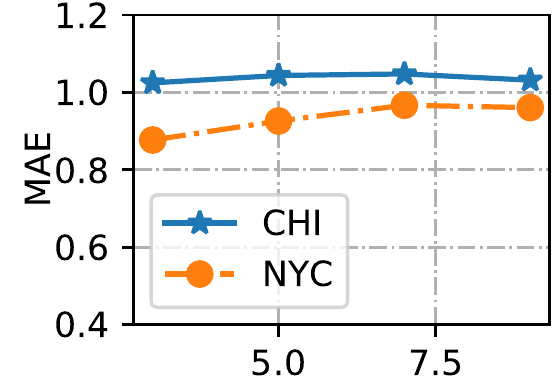}
    }
    \subfigure[\# Local Conv]{
        \centering
        \includegraphics[width=0.175\textwidth]{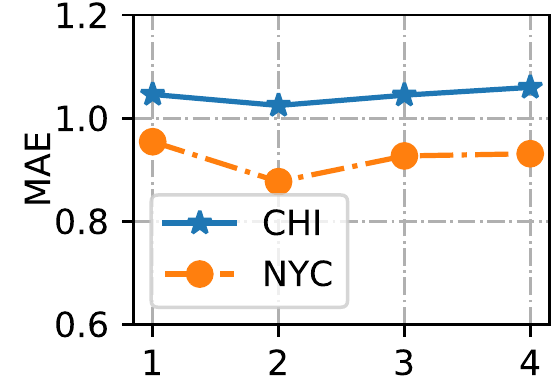}
    }
    \subfigure[\# Global Conv]{
        \centering
        \includegraphics[width=0.175\textwidth]{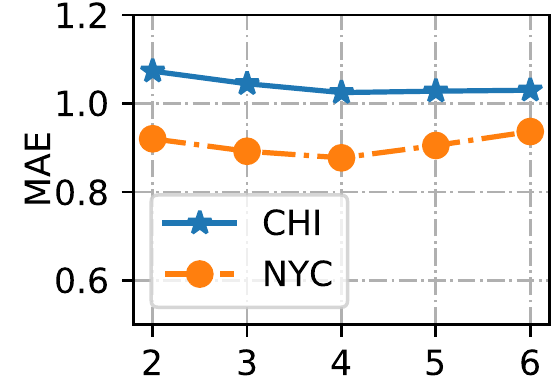}
    }
    \vspace{-0.05in}
    \caption{Impact study for various hyperparameters in \model's performance on Chicago and New York crime data, in terms of MAE and MAPE.}
    \label{fig:hyperparam}
\end{figure*}

\subsection{Model Robustness Study (RQ3)}
We also perform experiments to investigate the robustness of our \model\ method against data sparsity. To achieve this objective, we separately evaluate the prediction accuracy of regions with different density degrees. Here, the density degree of each region is estimated by the ratio of non-zero elements (crime occurs) in the region-specific crime occurrence sequence $\textbf{X}_r$. Specifically, we partition sparse regions with the crime density degree $\leq$ 0.5 into two groups (0.0, 0.25] and (0.25, 0.5]. The evaluation results are shown in Figure~\ref{fig:sparsity}.

We can observe that our \model\ consistently outperforms compared methods in all cases with different crime density degrees. This observation further validates the effectiveness of our \model\ framework in alleviating the data sparsity issue in crime data. The sparse supervision labels have negative impacts on the spatial and temporal relation learning with graph structures (\eg, GMAN, DMSTGCN), leading to sub-optimal prediction results of existing methods. This observation admits that current neural network-based spatial-temporal prediction methods (\eg, GNN-based or attention-based approaches) can hardly learn  high-quality representations for sparse regions. With the incorporation of our dual-stage self-supervised learning components into the graph-based crime prediction framework, \ie, hypergraph infomax network and local-global cross-view contrastive learning, we explore self-supervised signals from the crime data itself with augmented hypergraph learning tasks to enhance the model robustness in crime prediction. As such, this experiment again demonstrates that the spatial-temporal representation learning benefits greatly from our incorporated self-supervised signals, to offer accurate and robust crime forecasting performance.

\begin{figure*}
    \includegraphics[width=0.505\textwidth]{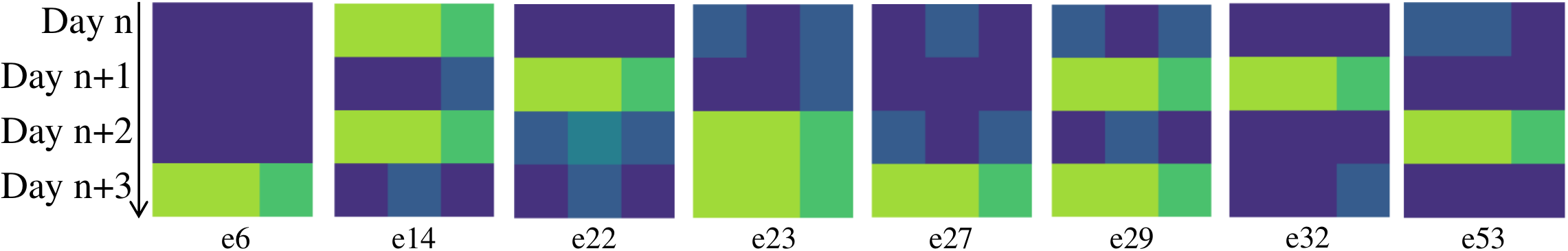}\quad
    \includegraphics[width=0.462\textwidth]{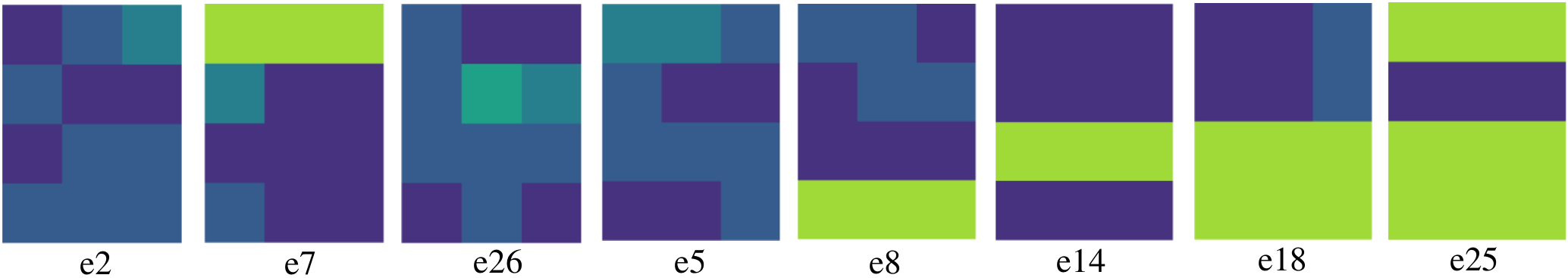}
    \vspace{0.05in}
    \includegraphics[width=0.524\textwidth]{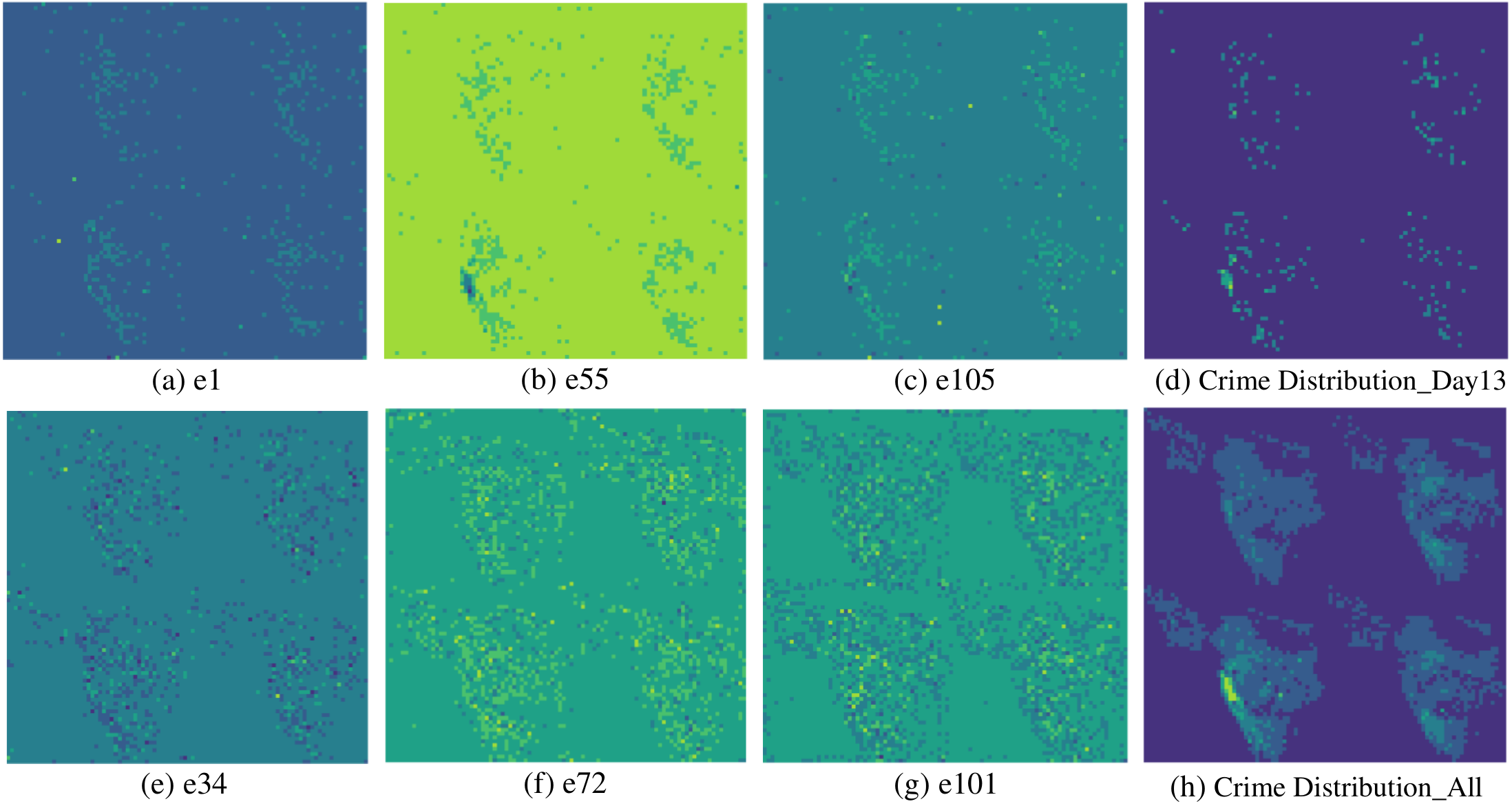}\quad
    \includegraphics[width=0.462\textwidth]{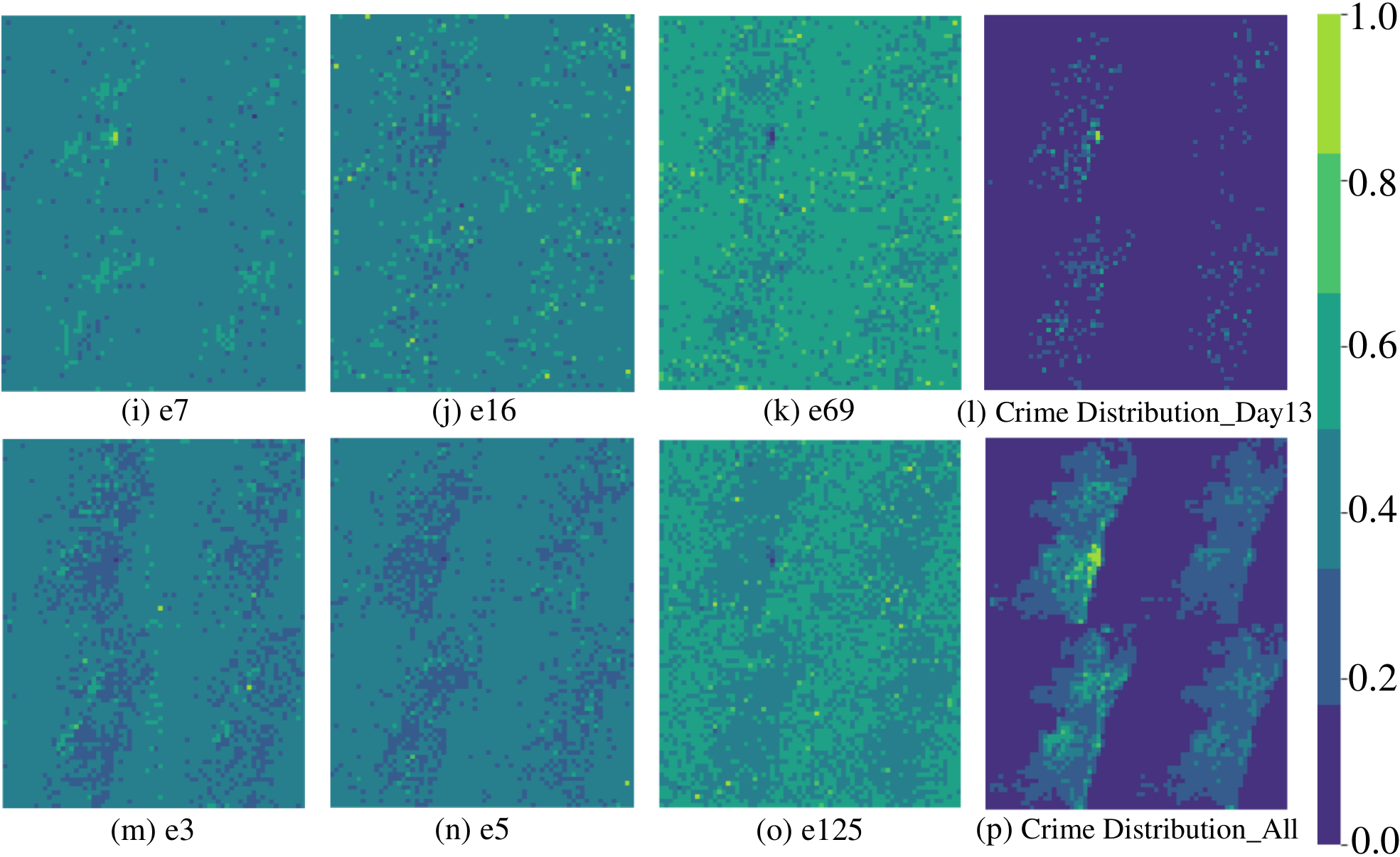}
    \vspace{0.05in}
    \includegraphics[width=0.228\textwidth]{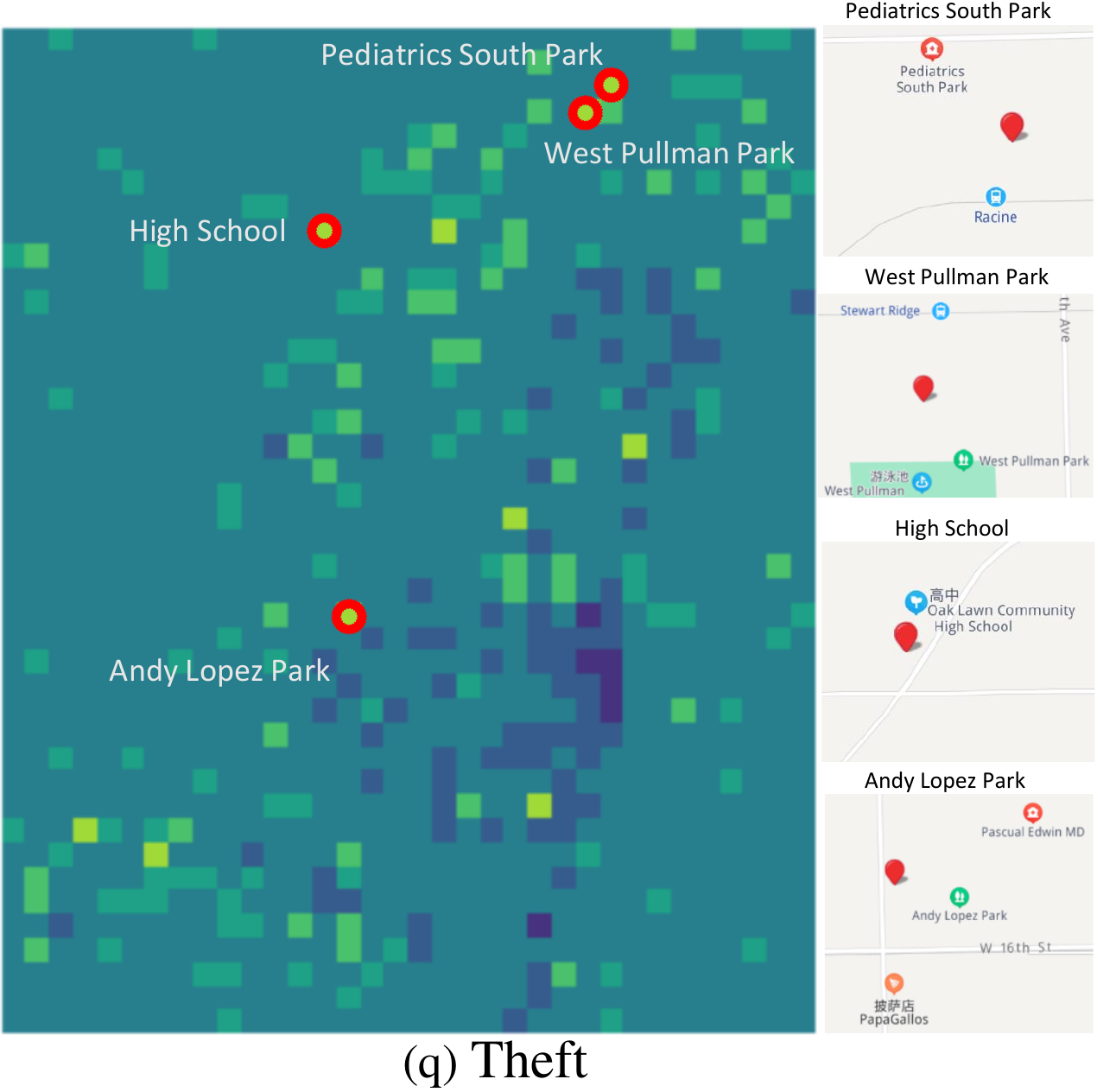}\quad
    \includegraphics[width=0.228\textwidth]{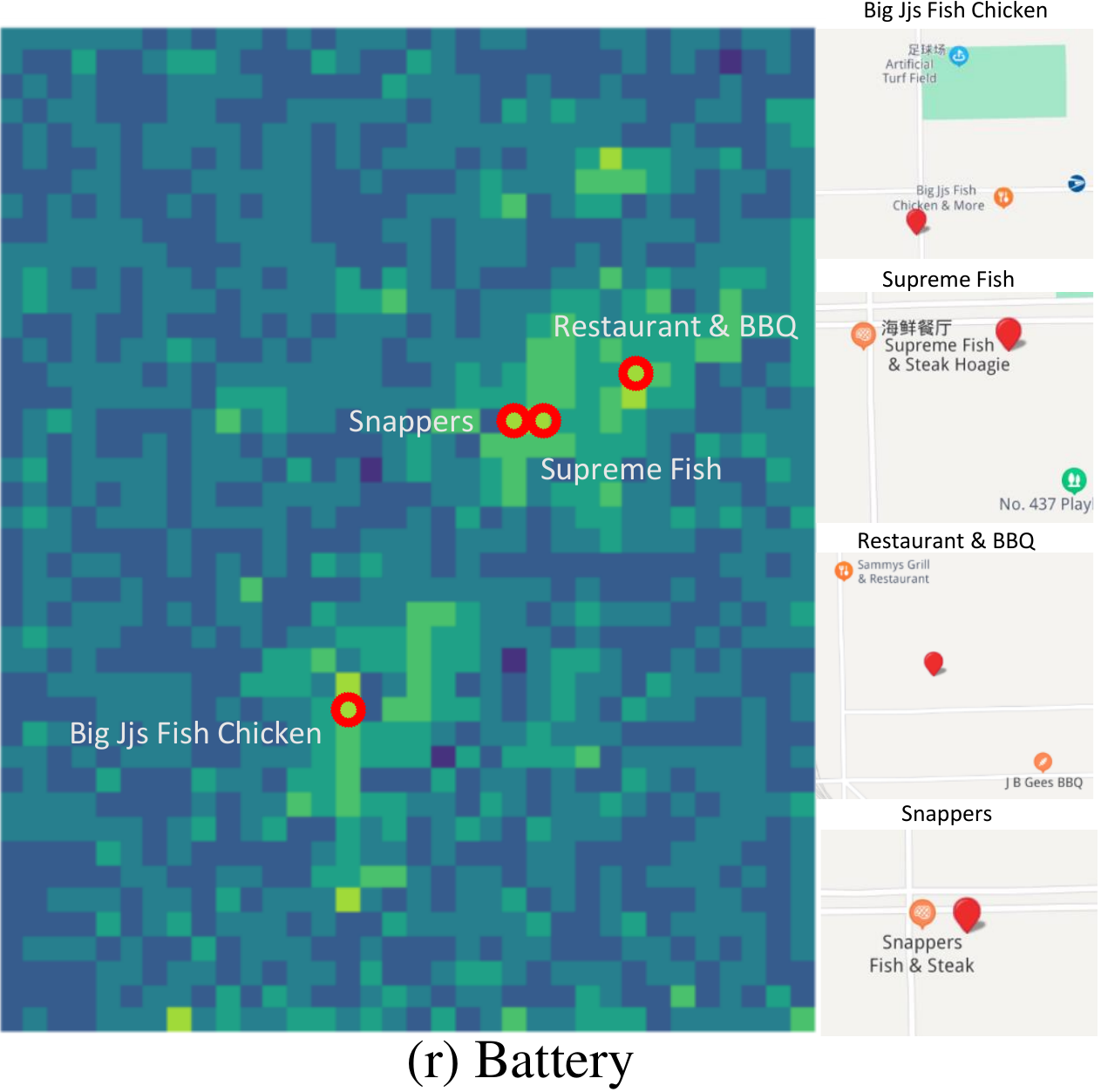}\quad
    \includegraphics[width=0.228\textwidth]{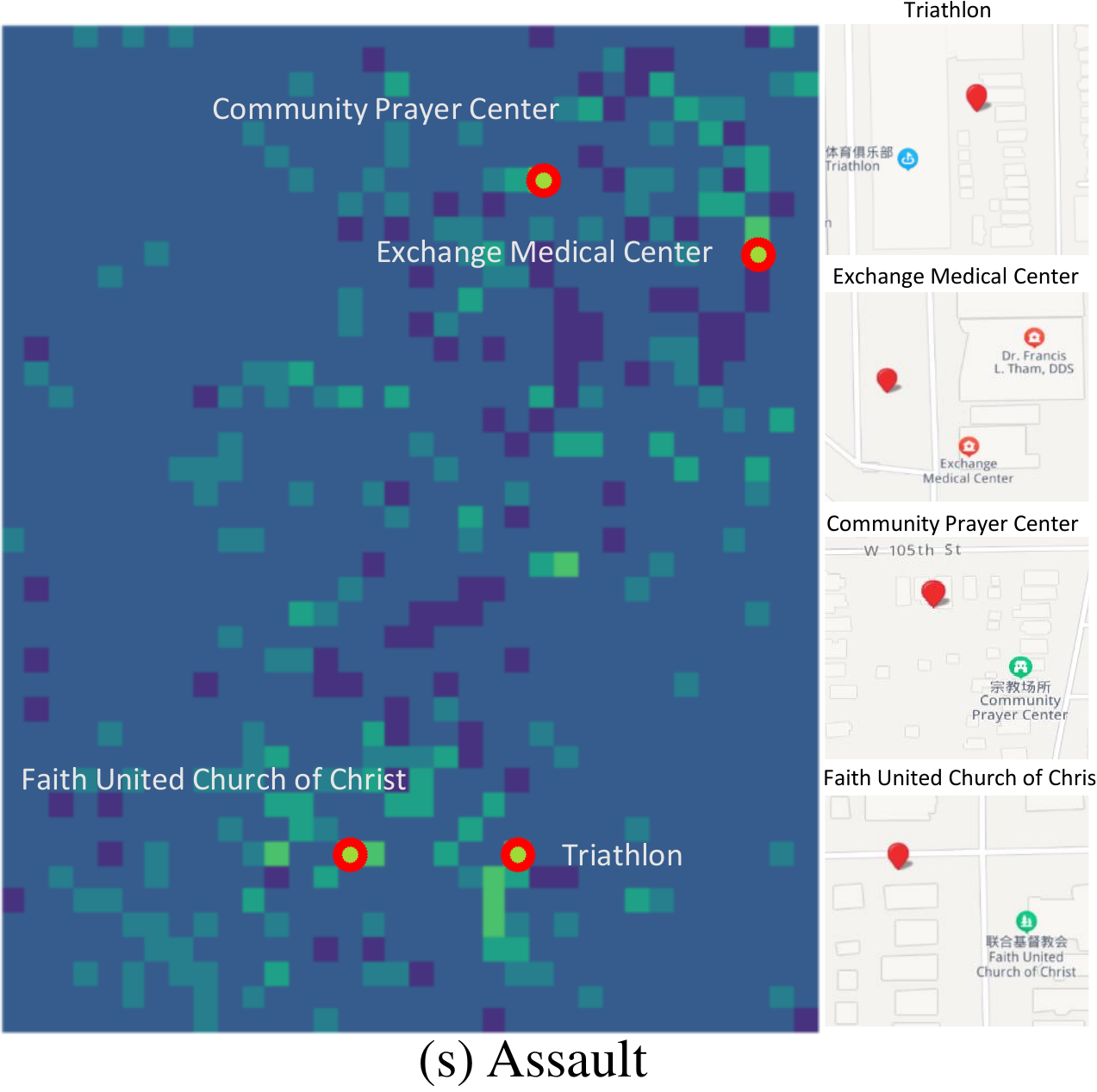}\quad
    \includegraphics[width=0.228\textwidth]{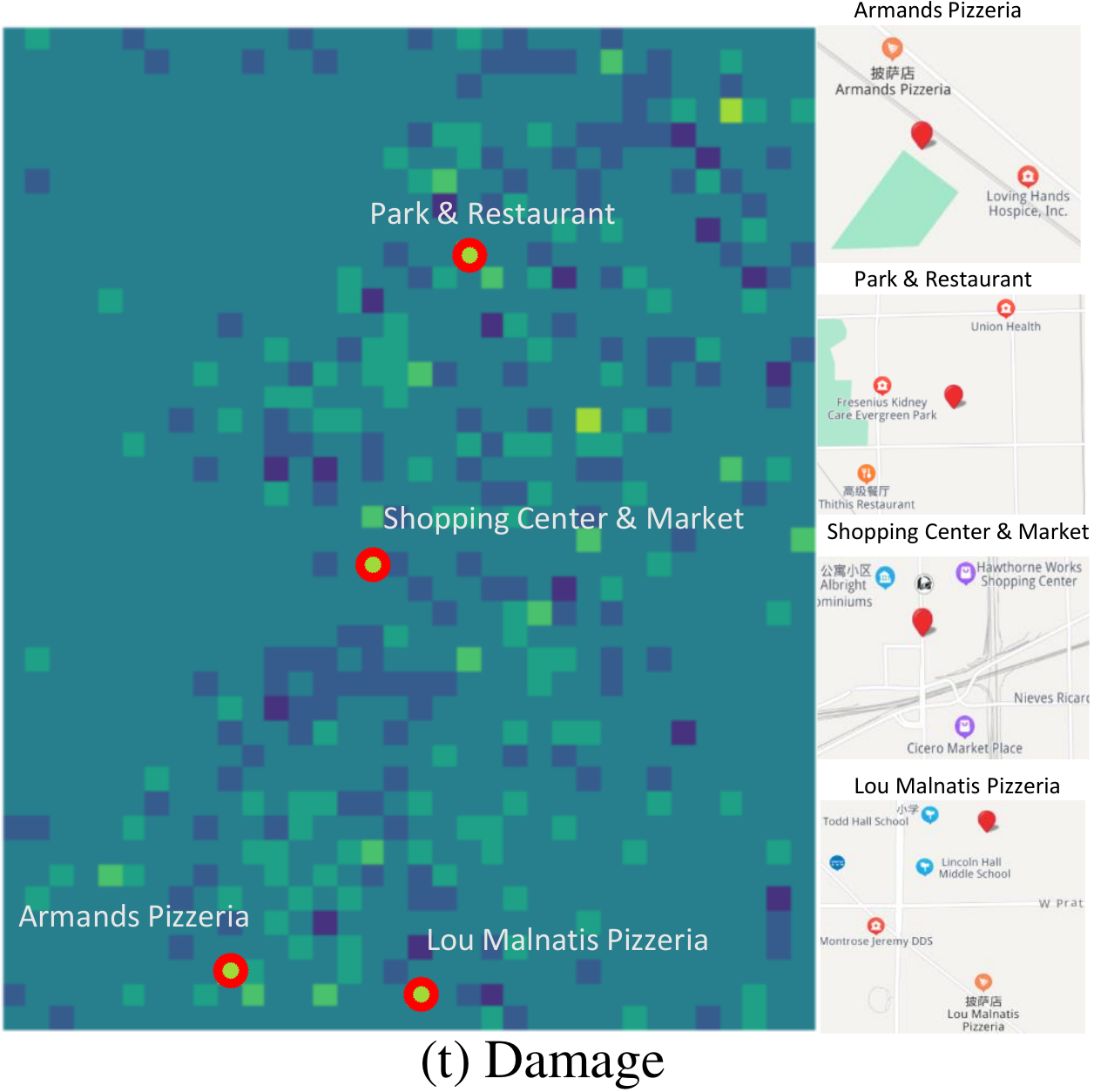}\quad
    \caption{Case study of our proposed \model\ framework. (i): 16 $4\times 3$ matrices in which each row corresponds to top-3 regions  with  the  highest  relevance  scores  (learned  from  our \model) to this hyperedge on a specific day. (ii): (a)-(p) sub-figures visualize the hyperedge-specific dependency scores over all regions in the global urban space. (iii) Highlighted several highly relevant geographical regions learned by our method for different types of crimes in Chicago. Those regions exhibit dependent relationships in terms of their region functionality from external source.}
    \vspace{-0.05in}
    \label{fig:case_study}
\end{figure*}

\subsection{Hyperparameter Studies (RQ4)}
To show the effect of different parameter settings, we perform experiments to evaluate the performance of our developed \emph{\model} framework with different configurations of key hyperparameters (\eg, \# of hypergraph channels, kernel size). When varying a specific hyperparameter for effect investigation, we keep other parameters with default values. The results are shown in Figure~\ref{fig:hyperparam}. We summarize the observations below to analyze the influence of different hyperparameters:

\begin{itemize}[leftmargin=*]

\item We vary the embedding dimensionality from the range of \{$2^2$, $2^3$, $2^4$, $2^5$\} and the best performance can be achieved with $d=16$. The prediction accuracy slightly degrades as we further increase the value of $d$, which suggests that larger embedding dimensionality does not always bring stronger model representation ability. The further increase of $d$ easily leads to the overfitting issue for spatial-temporal representations on sparse crime data. \\\vspace{-0.1in}

\item In our hypergraph structure learning module, hyperedges serve as latent representation channels for global region-wise relation modeling. We search the number of hyperedges $H$ in the range of \{$2^5$, $2^6$, $2^7$, $2^8$\}. The results indicate that $H=128$ is sufficient to well capture the global cross-region dependency with respect to their crime patterns.\\\vspace{-0.1in}

\item The best prediction accuracy is achieved with convolutional kernel size of 3. The larger size (\eg, 5,7,9) of convolution kernels may involve unexpected noise during the spatial-temporal information aggregation.

\end{itemize}

\subsection{Case Study (RQ5)}
We exploit the model interpretation ability of our \model\ in learning cross-region dependency with respect to the crime patterns under global context. Specifically, we first sample eight hyperedges (\eg, $e_{22}$, $e_{29}$, $e_{53}$) and generate a $4\times 3$ matrix for each one. In the matrix, each row represents top-3 regions with the highest relevance scores (learned from our \model) to this hyperedge on a specific day. For consistently, we apply the min-max normalization over the crime statistics and show the case studies in Figure~\ref{fig:case_study}. We can observe that those highly dependent regions indeed share similar crime patterns (shown with similar colors) across different time slots (\ie, days). This observation validates the effectiveness of our hypergraph dependency encoder in capturing global dependency among different geographical regions in a time-aware environment. Furthermore, we visualize the hyperedge-specific dependency scores over all regions in the global urban space. It can be seen that the encoded global region-specific crime patterns (shown in Figure~\ref{fig:case_study} (a)-(c), (e)-(g), (i)-(k), (m)-(o)) are consistent with ground-truth of crime occurrence (shown in Figure~\ref{fig:case_study} (d), (h), (l), (p)). This further confirms the effectiveness of our \model\ approach in capturing the complex and accurate cross-region dependencies to fit the real-world crime scenario in a city.

We further explore the explainability behind our prediction results with our \model\ framework. To be specific, we highlight several highly relevant geographical regions learned by our method for different types of crimes in Chicago. From these four sub-figures, we can notice that the learned highly dependent regions share similar functionality (\eg, city parks, restaurant zone, shopping center). Therefore, the learned hypergraph dependence structures between different regions, can show the learned insights leading to the crime prediction results by capturing the region-wise crime patterns.

\vspace{-0.05in}
\subsection{Model Efficiency Study (RQ6)}
We finally investigate the model scalability of our \model\ framework as compared to state-of-the-art spatial-temporal prediction techniques. All experiments are conducted with the default parameter settings in a computing sever with GTX 1080Ti and Inter Core i7-3770K. We study the efficiency of all compared methods by evaluating their running time of each training epoch on both NYC and Chicago datasets. Table~\ref{tab:training_time} lists the evaluation results. As can be seen, \model\ achieves better efficiency than most of baselines with respect to the model computational cost. The reason is that the designed hypergraph-guided self-supervised learning only incurs small computational cost with the augmented loss regularization. The high complexity of STDN lies in stacking many attention layers with explicit weight learning to fuse information.

\begin{table}[t]
    \centering
    \caption{Computational time cost investigation (seconds).}
    \label{tab:training_time}
    \begin{tabular}{crr||crr}
        \hline
        Model & NYC & CHI & Model & NYC & CHI\\
        \hline
        STGCN & 2.745 & 1.943 & DeepCrime & 12.926 & 11.550\\
        DMSTGCN  & 5.482 & 4.593 & ST-SHN & 17.872 & 16.310\\
        STtrans & 6.940 & 5.209 & DCRNN & 18.823 & 18.754 \\
        GMAN & 11.120 & 10.025 & STDN & 22.223 & 26.535\\
        ST-MetaNet & 11.938 & 11.100 & \underline{\emph{\model}} & \underline{12.355} & \underline{8.254}\\
        \hline
    \end{tabular}
    \vspace{-0.1in}
\end{table}
\section{Related Work}
\label{sec:relate}
In this section, we discuss the research work relevant to our studied urban crime prediction problem from four aspects: i) Deep spatial-temporal prediction techniques; ii) Graph neural networks for spatial-temporal data; iii) crime prediction; iv) self-supervised learning. 

\subsection{Deep Spatial-Temporal Prediction Techniques}
Many spatial-temporal prediction methods have been developed to model geographical and time-wise data patterns based on different neural networs~\cite{liu2020urban,2019deep}. For example, earlier predictive models are built on recurrent neural network for temporal information encoding, such as D-LSTM~\cite{yu2017deep} and ST-RNN~\cite{liu2016predicting}. In addition, there exist some hybrid spatial-temporal prediction methods, \eg, the convolutional shifted attention mechanism SDTN~\cite{yao2019revisiting}, recurrent attentive network MiST~\cite{huang2019mist}, and external factor fusion in UrbanFM~\cite{liang2019urbanfm}. ASPPA~\cite{zhao2020discovering} is a power-law attention mechanism which incorporates domain knowledge into the user mobility modeling with sub-sequential patterns. \cite{guo2021learning} aims to capture the dynamics and heterogeneity over spatial-temporal graph. Different from the above spatial-temporal prediction tasks, crime prediction presents unique challenge of data sparsity which cannot be easily handled by most existing prediction techniques. In this work, we propose a spatial-temporal self-supervised learning model with hypergraph neural architecture to capture dynamic crime patterns under data scarcity.


\subsection{Graph Neural Networks for Spatial-Temporal Data}
Graph neural network have shown its strong relational learning power in various applications~\cite{wu2020comprehensive,2021recent}, including fake news detection~\cite{zhang2020fakedetector}, knowledge graph learning~\cite{park2019estimating}, and recommender systems~\cite{2021knowledge}. In recent years, we have witnessed the development of research work applying graph neural networks to model spatial dependency among different locations~\cite{wang2020traffic}. For example, inspired by the effectiveness of graph convolutional network, spectral graph-based message passing methods have been developed to capture correlations among regions, \eg, STGCN~\cite{yubingspatio} and DCRNN~\cite{li2017diffusion}. To differentiate the embedding propagation between regions, graph attention network with spatial message passing has been proposed to aggregate features from correlated regions for traffic data modeling (\eg, GMAN~\cite{zheng2020gman}, ST-GDN~\cite{zhang2021traffic}), next location recommendation (\eg, STAN~\cite{luo2021stan}). Inspired by those work, we endow our spatial-temporal prediction framework with the global context-enhanced region dependency modeling under a self-supervised hypergraph learning paradigm.

\subsection{Crime Prediction}
There exist several relevant crime prediction models which explores the time-evolving crime occurrence patterns from both spatial and temporal dimensions~\cite{wang2016crime}. For example, Zhao~\etal~\cite{zhao2017modeling} formulates the crime prediction task as a tensor factorization architecture to learn the dependence among different geographical areas and time slots. DeepCrime~\cite{huang2018deepcrime} is the representative deep neural network model to integrate recurrent unit with attention mechanism to predict citywide crime occurrence. To investigate the fine-grained crime prediction, STtrans~\cite{2020hierarchically} develops a hierarchically structured Transformer network to fuse spatial-temporal-semantic relatedness from crime data. Despite the effectiveness of those methods, several issues remain less explored: i) the global cross-region crime dependence has been overlooked in existing crime prediction solutions; ii) the entire urban space usually involve sparse crime data, which undermines the representation capability of Transformer or attention networks. By effectively addressing these limitations, \model\ achieves better performance.

\subsection{Self-Supervised Learning}
Recently, self-supervised learning (SSL) becomes a promising solution to enhance the representation capability of neural networks, so as to address the limitation of high reliance on sufficient labeled data for model training~\cite{arsomngern2021self,liu2021self}. The effectiveness of SSL has been demonstrated in various domains, such as computer vision~\cite{kolesnikov2019revisiting}, nature language understanding~\cite{cheng2021self}, graph representation learning~\cite{tian2021self}. In self-supervised learning paradigms, models explore the supervision signals from the data itself with auxiliary learning tasks. Furthermore, contrastive-based SSL methods aim to reach agreement between generated correlated contrastive views~\cite{2022contrastive}. However, self-supervised learning is relatively less explored in spatial-temporal data prediction. This work brings SSL's superiority into the urban crime prediction task to tackle the challenges of crime data sparsity and skewed distribution. Towards this end, the proposed \model\ prediction framework integrates the generative and contrastive learning with a dual-stage self-supervised augmentation scheme for jointly modeling local and global crime patterns.


\section{Conclusion}
\label{sec:conclusion}

In this paper, we introduce the hypergraph contrastive learning into the crime prediction by proposing a new spatial-temporal self-supervised learning framework \model. The proposed \model\ first augments the local spatial-temporal encoder with hypergraph-based global relation learning. Then, on the hypergraph structure, we introduce a dual-stage self-supervised learning paradigm to enhance the representation ability of \model\ with sparse supervision crime signals. Extensive empirical results on real-life urban crime datasets demonstrate the effectiveness of our \model\ learning method. 



\section*{Acknowledgments}
We thank the anonymous reviewers for their constructive feedback and comments. This work is supported in part by National Nature Science Foundation of China (62072188), Major Project of National Social Science Foundation of China (18ZDA062), Science and Technology Program of Guangdong Province (2019A050510010).

\clearpage

\bibliographystyle{abbrv}
\bibliography{refs} 

\end{document}